\documentclass[5p,times]{elsarticle}

\usepackage{amssymb}
\usepackage{svg}
\usepackage{caption}
\usepackage{subcaption}
\usepackage{tabularray}
\usepackage{url}
\usepackage[all]{nowidow}
\usepackage{xcolor}
\usepackage{framed}

\usepackage[linesnumbered,ruled,vlined]{algorithm2e}

\usepackage{enumitem}
\setlist[description]{leftmargin=0.5cm}

\journal{Future Generation Computer Systems}

\begin{document}

\begin{frontmatter}



\title{A Framework for testing Federated Learning algorithms using an edge-like environment}


\author[inst1]{Felipe Machado Schwanck}

\affiliation[inst1]{organization={Federal University of Rio Grande do Sul (UFRGS), Institute of Informatics},
            addressline={Av. Bento Gonçalvez, 9500}, 
            city={Porto Alegre},
            postcode={91.509-900}, 
            state={RS},
            country={Brazil}}
\author[inst1]{Marcos Tomazzoli Leipnitz}
\author[inst1]{Joel Luís Carbonera}
\author[inst1]{Juliano Araujo Wickboldt}

\begin{abstract}
Federated Learning (FL) is a machine learning paradigm in which many clients cooperatively train a single centralized model while keeping their data private and decentralized. FL is commonly used in edge computing, which involves placing computer workloads (both hardware and software) as close as possible to the edge, where data are created and where actions are occurring, enabling faster response times, greater data privacy, and reduced data transfer costs. However, due to the heterogeneous data distributions/contents of clients, it is non-trivial to accurately evaluate the contributions of local models in global centralized model aggregation. This is an example of a major challenge in FL, commonly known as data imbalance or class imbalance. In general, testing and evaluating FL algorithms can be a very difficult and complex task due to the distributed nature of the systems. In this work, a framework is proposed and implemented to evaluate FL algorithms in a more easy and scalable way. This framework is evaluated over a distributed edge-like environment managed by a container orchestration platform (i.e. Kubernetes).
\end{abstract}



\begin{keyword}
Federated Learning \sep Edge Computing \sep Kubernetes \sep Microservices \sep Development Framework
\end{keyword}

\end{frontmatter}




\section{Introduction}
\label{sec:intro}

Federated Learning (FL) is a machine learning solution designed to train machine learning models while keeping the data private and decentralized \cite{fl-fed-avg}. The main idea of FL is for each client to train its local model using its data and, afterward, upload the generated local model to a single centralized server, where the regional models of the participant clients will be aggregated and weighted to create a global model. FL is commonly used in edge computing, which involves placing computer workloads (both hardware and software) as close as possible to the edge, where the data is being created and where actions are occurring, enabling faster response times, greater data privacy, and reduced data transfer costs \cite{edge-computing-ibm}. Thus, FL can be used with a variety of clients, such as smartphones, sensors, IoT devices, and data silos (distributed databases that need to keep their data private from the outside world). However, as seen in \cite{fl-imbalance-astrea}, in general, the data distribution of the mobile systems and other similar settings is imbalanced, which can increase the bias of the model and impact in a negative way its performance. Different approaches have been proposed, such as Deep Reinforcement Learning (DRL) \cite{drl-book}, as a solution for this problem.

Although regular centralized machine learning may outperform FL in prediction performance \cite{fl-performance}, the entire dataset must be shared. Its first application was in Google GBoard \cite{fl-keyboard}, which learns from every smartphone using Gboard without sharing user data. Since then, FL applicability has advanced to various fields such as autonomous vehicles, traffic prediction and monitoring, telecom, IoT, pharmaceutics, industrial management, industrial IoT, and healthcare and medical AI \cite{fl-applications}. Moreover, the number of academic publications with FL as main subject has increased significantly since its conception~\cite{fl-fed-avg}.

The increase in IoT devices has greatly enabled FL. The total number of IoT connections will reach 83 billion by 2024, rising from 35 billion connections in 2020, a growth of 130\% over the next four years \cite{barker2020internet}. The industrial sector has been identified as a critical driver of this growth. Expansion will be driven by the increasing use of private networks that leverage cellular network standards \cite{edge-iot-devices}. The evolution of the IoT device's computational power has also enabled FL; since an edge computer can process data locally, its sensors (e.g. cameras) could collect samples (e.g. images or frames) at a higher resolution and a higher frequency (such as frame rate) than would be possible if the data had to be sent to the cloud for processing \cite{edge-vs-iot-ibm}.

One of the many challenges that have come up with the advance of FL is dealing with data imbalance and heterogeneity, as seen in \cite{fl-applications}. In FL, using their local data, each edge node trains a shared model. As a result, data distribution from those edge devices is based on their many uses. Imagine a scenario, for example, of the distribution of cameras in a surveillance system. Compared to cameras located in the wild, cameras in the park, for example, capture more photographs of humans. Also, the size of the dataset that each one of those cameras will have to train their local models might differ by a large magnitude since a park might have much more data to input than a camera in the wild. Furthermore, an approach that has been proposed in many studies to dynamically address weight for the local models of clients participating in the FL global model and, therefore, deal with the heterogeneous data is DRL \cite{fl-drl-adaptive, fl-drl-dear-fsac, fl-drl-fair, fl-drl-energy-consumption}. 

DRL has gathered much attention recently. Recent studies have shown impressive results in activities as diverse as autonomous driving \cite{drl-autonomous-driving}, game playing \cite{drl-starcraft}, molecular recombination \cite{drl-molecular-remodeling}, and robotics \cite{drl-robotics-walk}. In all those applications, it has been used in computer programs to teach them how to solve complex problems, for example, how to fly model helicopters and perform aerobatic maneuvers, and, in some applications, it has already outsmarted some of the most skilled humans, such as in Atari, Go, poker and StarCraft.

Testing and evaluating FL algorithms can be a very difficult task to accomplish. FL intrinsically creates complex distributed systems with non-trivial interactions among its participants. Therefore, this work proposes and implements a framework for testing FL algorithms that enables users to easily create different training scenarios by simply changing configuration parameters. These include computing and data distributions, datasets, global model aggregations, local client models, and server/client training parameters. The framework is also capable of collecting and visualizing training results and resource usage metrics (e.g., CPU, memory.). Experiments have been conducted over a realistic edge-like environment managed by a Kubernetes container orchestration platform, with varied parameters on top of a proof-of-concept (PoC) implementation of the proposed framework to demonstrate its capabilities. 


The remainder of this work is organized as follows. Section~\ref {chap-literature-review} presents a literature review including edge computing and FL concepts, as well as a brief review of FL testing and development proposals. Section~\ref{chap-solution} introduces the conceptual architecture of the proposed framework. Section~\ref{chap-tools-framework} discusses the tools and frameworks used to implement the PoC solution. Section~\ref{chap-poc} details the PoC developed. Sections~\ref{section-experimental-setup},~\ref{section-results}, and~\ref{section-monitoring} present the experimental setup, the results obtained, and how resource usage and network traffic can be monitored, respectively. Finally, Section~\ref{section-conclusion} concludes the work with final remarks and a perspective on future work.
\section{Literature Review}\label{chap-literature-review}

This section presents the main background concepts used in this work and provides an overview of the fundamentals in edge computing, FL, as well as the current state-of-the-art in developing and testing approaches and frameworks available to implement FL solutions.

\subsection{Edge Computing}

Data is increasingly produced at the edge of the network; therefore, processing the data at the network's edge would be more efficient. With the advancement of telecommunication services and the increase of the necessity for low-latency computing, the edge computing paradigm has been motivated. In edge computing, instead of having computer workloads (both hardware and software) centralized in a data center (cloud), we have them as close as possible to the edge, where the data is being created and where actions are occurring, thus benefiting lower latency, greater data privacy, and reduced data transfer costs. For \cite{edge-vision-challenges}, edge computing refers to the enabling technologies allowing computation to be performed at the edge of the network, on downstream data on behalf of cloud services, and upstream data on behalf of IoT services. Therefore, edge devices can be any device with Internet access, such as smartphones, smart cars, or other IoT devices.

Multi-access Edge Computing (MEC) \cite{edge-mec} is proposed as a critical solution that enables operators to open their networks to new services and IT ecosystems and leverage edge-cloud benefits in their networks and systems since it places storage and computation at the network edge. The proximity of the end users and connected devices provides low latency and high bandwidth while minimizing centralized cloud limitations such as delay, access bottlenecks, and single points of failure. 


MEC use cases can be seen in real-time traffic monitoring \cite{edge-realtime-traffic} and autonomous vehicles \cite{edge-vehicles}. In traffic monitoring, real-time and accurate video analysis is critical and challenging work, especially in situations with complex street scenes; therefore, edge computing-based video pre-processing is proposed to eliminate the redundant frames edge devices need to process since a considerable amount of vehicle video data is generated. Also, the decentralized and highly available nature of multi-access edge computing is taken advantage of to collect, store, and analyze city traffic data in multiple sensors. For autonomous vehicles, a large amount of real-time data processing from different sensors at high speed is needed to guarantee driver safety.

\subsection{Federated Learning}

Federated Learning (FL) is a distributed form of machine learning proposed by Google \cite{fl-fed-avg} to train models at scale while allowing the user data to be private. In Federated Averaging (FedAvg), the server aggregates the model updates using simple averaging. It returns the new model parameters to the client devices, which continue training using the updated model parameters. Google's proposal provided the first definition of FL, as well as the Federated Optimization \cite{fl-optimization} approach to improve these federated algorithms further. Advanced Federated Optimization \cite{fl-fedopt} is a variant of the Stochastic Gradient Descent (SGD) algorithm commonly used in centralized training. In FedOpt, each local node applies SGD to its local data to compute the gradients and then sends them to a central server. The server then aggregates the gradients from all the nodes to update the global model. Adaptive Federated Optimization with Yogi (FedYogi) incorporates a momentum-based optimizer called Yogi after the central server aggregates the model updates using the FedAvg algorithm to improve the convergence rate. Federated Averaging with Momentum \cite{fl-fedavgm} incorporates a momentum-based optimizer, similar to the Yogi optimizer in FedYogi. The critical difference is that it uses a combination of the gradients from the current iteration and the gradients from the previous iteration to update the model parameters. This allows the optimizer to maintain a direction of movement even when the gradient changes direction, which helps to smooth out noisy gradients and accelerate convergence. In the ``Federated Learning: Collaborative Machine Learning without Centralized Training Data" blog post \cite{google-blog}, Google explains how FL is enabling mobile phones to collaboratively learn a shared prediction model while keeping all the training data on the device, decoupling the ability to do machine learning from the need to store the data in the cloud. In addition, the current use of FL to predict keyboard words in Google's Gboard \cite{fl-keyboard} and how it can be used for photo ranking and further improving language models. 

FL usually deals with data distributed across multiple devices. In such settings, data is usually non-independently and identically distributed (i.e., non-IID). One of the main challenges in FL is dealing with the heterogeneity of the data distribution among the parties since data distribution from those edge devices is based on their many uses \cite{fl-applications}.

Furthermore, many use cases in FL have data samples distributed among multiple devices, which are not always synchronized and may have limited connectivity. Thus, it cannot train these devices in parallel and directly aggregate them, as it cannot guarantee device availability or data homogeneity. Understanding how to properly select clients and weigh each client's contributions in the global model remains an open problem in FL. An example of a practical scenario of data imbalance and heterogeneity in which DRL was used in FL as a solution can be seen in recent work proposed to deal with blade icing detection in distributed wind turbines \cite{fl-class-imbalance}. Wind turbines closer to the sea experience windy and snowy weather, while those closer to the continent deal with windy and rainy conditions. This heterogeneity introduces a bias in the local models since one might be more susceptible to icing than another. Therefore, since our objective is to identify icing, clients who experience more icing should have a different weight assigned to their contribution to the global model than others without. 

\subsection{Federated Learning Testing and Development}

FL has gained traction as a method for training machine learning models across decentralized data sources while ensuring privacy. Notable frameworks for development and testing of FL solutions have recently been introduced, some of which consider cloud and edge infrastructures for deployment and experimentation \cite{Duan2023FL-Edge-Survey}. One example is Flower, which offers a solution for research and large-scale experimentation on FL algorithms and simulation of system-level challenges, such as CPU and memory limitation of FL client devices~\cite{fl-flower}. Flower stands out among previous FL implementations because it includes important features, such as multi-node experimentation, client heterogeneity, and language agnosticity, in comparison to other frameworks like TFF \cite{TFF}, Syft \cite{fl-pysyft}, FedScale \cite{lai2022fedscale}, FedProx \cite{fl-fedprox}, and LEAF \cite{fl-leaf}. We actually build on top of Flower to develop the framework presented in this article, as further detailed in Sections \ref{chap-tools-framework} and \ref{chap-poc}. 

FedML is a FL implementation library that provides a standardized set of benchmarks for algorithms, models, and datasets \cite{fl-fedml}. It also enables developers to use Python to customize experiments to run on IoT devices (e.g., Raspberry Pi or NVidia Jetson platforms). 
FATE is a FL development framework that claims to be production-oriented, as opposed to most others that are research-oriented. The platform was originally developed and open-sourced by Webank’s AI Department and supports industrial applications with robust data protection \cite{fl-fate}. The maintainers also release a set of scripts and configuration files to allow the deployment of FATE on top of Kubernetes, which is called KubeFATE. With the same main objective to provide enhanced privacy preservation for the distributed deployment of FL applications, Parra-Ullauri et al. proposed kubeFlower \cite{PARRAULLAURI2024558}. The authors introduce a unique privacy preservation approach and show that kubeFlower outperforms kubeFATE in terms of time to deploy. 

Together, these frameworks significantly advance the development and application of FL across various domains. However, none of them focused specifically on simplifying the creation of different training scenarios with simple configuration parameters, including computing and data distributions, datasets, models, and server/client training parameters. Our framework not only targets this challenge, but also automates the collection and visualization of training results and resource usage metrics, making the job of the FL developer/experimenter much easier.



\section{Conceptual Framework}\label{chap-solution}



\begin{figure*}[ht]
\centering
\includegraphics[width=0.9\textwidth]{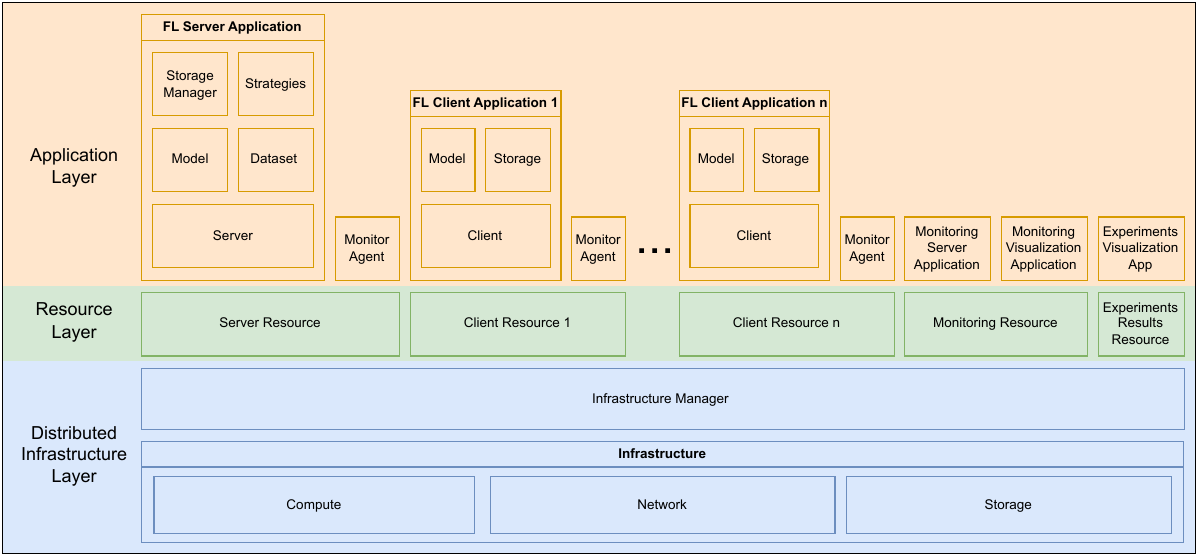}
\caption{High-level overview of the conceptual framework proposed}
\label{fig:concept-architecture-fig}
\end{figure*}

The main objective of the proposed framework is to enable FL testing in a platform where users can easily change parameters to create different scenarios in a distributed computing environment. These include different computing and data distributions, datasets, global model aggregations, client models, and server and client training parameters. The framework should also be capable of collecting training results and resource usage metrics. Figure \ref{fig:concept-architecture-fig} presents an overview of the proposed framework's main components in a conceptual architecture. As can be seen, the framework is designed in three independent layers to enable easier development and segregation of functions so that if improvements are made in one layer, they do not affect the others. The following subsections detail each layers and its components. 

\subsection{Distributed Infrastructure Layer}\label{sec-concept-infrastructure}
The Distributed Infrastructure Layer is the framework's bottom layer responsible for all the infrastructure needed to create a distributed computing environment. It scales computing, storage, and networking from a pool of resources to meet user input parameters. These parameters, which include declarative commands for the infrastructure layer to provision and manage the resources needed to create the testing scenario, can be defined by the user using configuration files typically written in YAML or JSON. The infrastructure manager then ensures that the resources are provisioned and maintained per the user's desired state, providing idempotence and enabling Infrastructure-as-Code (IaC) practices for tracking, reviewing, and auditing infrastructure changes.

For example, consider a user who wants to configure a scenario of FL using ten clients, whereas each client has a specific computing requirement of four CPU cores and 4GB of RAM per client. The distributed infrastructure manager will then configure the resource layer from its pool of resources (computing, storage, and networking) with ten clients, each with the specified specifications to meet user requirements. Having an infrastructure layer independent from the resource and application layers eases the usage of different computational devices, since their provisioning is transparent for the application. This independence is fundamental for configuring different testing scenarios where other edge computing devices can be used as clients without the user having to deal with infrastructure provision directly.


\subsection{Resource Layer}\label{sub-concept-resource}

The resource layer, positioned on top of the distributed infrastructure layer, plays a crucial role in the framework. As mentioned, it matches the user's configuration for the FL training scenario. We can categorize the distribution of resources into four major categories: server resources, client resources, monitoring resources, and experiment results resources. The server resources contains all the necessary resources to run the FL server, matching the user configuration, and the same can be said for the client resources. On the other hand, the monitoring resources includes all the resources needed to run the monitoring server, which monitors the distributed infrastructure layer and records the usage of CPU, RAM, and network for all deployed resources. Finally, the experiment results resources include all the resources necessary to run an application responsible for displaying the results of the experiments run in the framework.

\subsection{Application Layer}\label{sec-concept-application}

The application layer runs on top of the resource layer provisioned by the distributed infrastructure. It contains all the applications necessary to run an FL algorithm in the framework, such as the FL server application, FL client application, monitor agent application, monitor server application, and the experiment results visualization application. Subsections \ref{sub-concept-server-application} to \ref{sub-concept-experiment-visualization} detail the application layer components.

\subsubsection{FL Server Application}\label{sub-concept-server-application}

The FL Server application can be divided into five major components that enable FL testing in a distributed environment.

\begin{description}
    \item[Server] is responsible for communicating with the clients and controlling the entire FL learning process, which includes selecting available clients to start training; control of the number of FL server rounds and round timeout; global model parameters, aggregation and distribution, and model retrieval parameters.
    
    \item[Model] is responsible for the server-side initialization of the global model parameters since some global model aggregation strategies need it to start FL learning training and enable the usage of different models to learn weight balancing of client models to distribute parameters. 
    
    \item[Strategies] selects and configures supported global model aggregation strategies from the user-specified configuration.
    
    \item[Dataset] is a collection of datasets supported by the framework used by the application to handle the dataset configuration in memory and to properly obtain the raw data to create data distributions via Storage Manager.
    
    \item[Storage Manager] is responsible for distributing the raw data of the configured dataset throughout the distributed storage infrastructure. For example, suppose a user wants to test an FL algorithm in a dataset using an unbalanced non-iid data distribution with ten clients. The storage manager will separate the data into ten unbalanced and biased non-iid data parts. The storage manager of the server application is also responsible for managing where each experiment will write its data and where the client will output their results data in the system.

\end{description}

\subsubsection{FL Client Application}\label{sub-concept-client-application}

The FL client application can be divided into three major components.

    \textbf{Client:} is responsible for the training and testing algorithms run on the client side and the connection with the server.
    
    \textbf{Model:} is responsible for selecting the desired model configured by the user to be used by the client to train it.
    
    \textbf{Storage Manager:} is responsible for loading the distributed data for training in the client and handling the experiments and test results of the client in the storage.


\subsubsection{Monitoring Application}\label{sub-concept-monitoring}

The Monitoring Server Application is responsible for gathering resource usage data of the distributed infrastructure and storing it for further visualization and analysis by the Monitoring Visualization Application. Each Monitoring Agent is responsible for gathering local resource usage data from the distributed resources and sending the metrics to the Monitoring Server Application.

\subsubsection{Experiment Visualization Application}\label{sub-concept-experiment-visualization}

This application is responsible for querying data from already run experiments and enabling the user to visualize the data for each step of the FL training experiment in customizable dashboards.

\section{Tools and Frameworks}\label{chap-tools-framework}

This section provides an overview of the tools and frameworks considered to implement the conceptual framework described in Section~\ref{chap-solution}. 

\subsection{Containerization and Orchestration Technologies}\label{sec-tools-orchestration}

This section presents tools for building, deploying, and orchestrating containers in distributed computing environments.  

\subsubsection{Docker}\label{sub-tools-docker}


Docker\footnote{https://docker.com/ (accessed April 24th, 2024)} is the most popular technology for simplifying packaging software and its dependencies into a loosely isolated environment called a container. Containers provide a robust solution for bundling software and its dependencies into a transportable unit that can run seamlessly across diverse computing environments. A containerized application encapsulates all its libraries, settings, and tools within an isolated environment,  enforcing security and allowing many containers to run simultaneously on a single. Unlike virtual machines, containers share the host operating system's kernel and libraries, resulting in lower resource usage and faster start-up times. The inherent portability of containers also eliminates the need to reconfigure applications for distinct environments, ensuring uniformity and dependability. Thus, developers can avoid the burden of addressing compatibility issues with varied hardware and software and concentrate solely on the application's functionality.

Although we used Docker as the container runtime to implement the framework's application layer, other options that comply with the Open Container Initiative (OCI) standards could be used. One example is Containerd\footnote{https://containerd.io/ (accessed April 24th, 2024)}, originally part of Docker and now a standalone project under the Cloud Native Computing Foundation (CNCF) focusing on high-performance container lifecycle management. CRI-O\footnote{https://cri-o.io/ (accessed April 24th, 2024)} is another lightweight, Kubernetes-specific alternative optimized for security and efficiency in resource-constrained environments like edge computing. Podman is a Docker-compatible solution that could also be considered for environments requiring rootless or daemon-less container execution for increased security. Therefore, the framework's application layer remains adaptable for various on-premise, cloud, and edge deployment scenarios, as it is not tied to specific containerization technologies.


\subsubsection{Kubernetes}\label{sub-tools-kubernetes}
Kubernetes\footnote{https://kubernetes.io/ (accessed April 24th, 2024)} is a portable, extensible, open-source container orchestration platform for managing containerized workloads and services via declarative configuration and automation. It has a large, rapidly growing ecosystem maintained by the CNCF. Kubernetes services, support, and tools are widely available, enabling container deployment, scaling, and management automation in distributed computing environments. A Kubernetes cluster consists of worker machines called nodes, which run containerized applications. Every cluster has at least one worker node. The worker node(s) host the Pods, the smallest deployment unit representing a single instance of a running process. A Pod describes a component of the application workload and encapsulates one or more containers, storage resources, a unique IP address, and other configuration options. Heterogeneous edge computing devices can be easily integrated and managed by Kubernetes as worker nodes, facilitating the deployment of different FL algorithms and the creation of varying testing scenarios in a distributed infrastructure environment deployed on-premise or in the cloud.

In addition, different Kubernetes flavors offer varied levels of flexibility, especially in edge or resource-constrained environments. K3s\footnote{https://k3s.io/ (accessed October 4th, 2024)}, for example, is a lightweight Kubernetes distribution ideal for edge computing and IoT, as it reduces resource overhead by removing non-essential features, making it suitable for smaller clusters with limited hardware. MicroK8s\footnote{https://microk8s.io/ (accessed October 4th, 2024)} is another Kubernetes variant designed for easy deployment, offering a single-command setup and modular features, allowing users to customize it based on their needs while retaining core Kubernetes functionalities. For edge-specific scenarios, KubeEdge\footnote{https://kubeedge.io/ (accessed October 4th, 2024)} extends the Kubernetes capabilities to allow seamless integration between Kubernetes clusters and edge devices, enabling real-time processing closer to the data source. It supports scenarios where connectivity is intermittent, or data privacy is critical by ensuring that edge nodes can independently function even when disconnected from the cluster. These alternatives, from traditional Kubernetes clusters to lightweight or edge-native solutions like KubeEdge, offer flexible cloud-agnostic deployments for the framework's distributed infrastructure layer tailored to user requirements.

In this work, we used an on-premise installation of Kubernetes for the distributed infrastructure layer, but cloud-based container management services offered by cloud providers, like AWS EKS\footnote{https://aws.amazon.com/eks/ (accessed October 4th, 2024)}, Google GKE\footnote{https://cloud.google.com/kubernetes-engine/?hl=en (accessed October 4th, 2024)}, or Azure AKS\footnote{https://azure.microsoft.com/en-us/products/kubernetes-service (accessed October 4th, 2024)}, can also be considered to offer scalable, fully managed Kubernetes alternatives. In conjunction with edge services like AWS IoT Greengrass, Google Cloud IoT Edge, or Azure IoT Edge, these solutions can handle server-side orchestration and model aggregation in the cloud while supporting machine learning model deployment, IoT device management, and serverless functions at the edge to train models locally, reducing operational overhead and integrating seamlessly with existing cloud ecosystems. Additionally, such services can dynamically scale resources on demand, making them ideal for fluctuating workloads, such as in FL. Therefore, cloud-based solutions can be explored to enhance the framework's scalability and flexibility, especially for organizations already using cloud ecosystems.

\subsection{Libraries and Frameworks} \label{sec-tools-libraries}

This section approaches the libraries and frameworks we employed to implement and execute FL algorithms.

\subsubsection{PyTorch}\label{sub-tools-pytorch}

PyTorch\footnote{https://pytorch.org/ (accessed April 24th, 2024)} is a popular open-source machine learning library that was created by Meta's AI research team. It is used to develop and train deep learning models and is written in Python, which makes it easy to use and integrate with other Python libraries.
PyTorch was the main library for training and testing the FL models used in this work.

\subsubsection{Poetry}\label{sub-tools-poetry}

Poetry\footnote{https://python-poetry.org/ (accessed April 24th, 2024)} is a tool for dependency management and packaging in Python. It allows you to declare the libraries your project depends on, and it will manage (install/update) them for you. Poetry offers a lock file to ensure repeatable installs and can build your project for distribution. It was used in this work to create the packages of the FL client and server applications.

\subsubsection{Flower}\label{sub-tools-flower}

As discussed in this work, the concept of FL emerged in response to the need to leverage data from multiple devices while ensuring their privacy. However, as opposed to traditional machine learning, FL introduces two additional challenges: scaling to various clients and dealing with data heterogeneity.


To address these challenges, as proposed in \cite{fl-flower}, Flower has been developed as an open-source framework for building FL systems. Flower provides two primary interfaces: the client and the server. These interfaces enable the decentralization of standard centralized machine learning solutions by implementing the necessary methods, making building and deploying FL systems easier. In Flower's architecture\footnote{https://flower.dev/docs/architecture.html (accessed April 4th, 2024)} each edge device in the training process runs a Flower client containing a local machine learning model. Flower provides a transparent connection via the Edge Client Proxy using an RPC protocol, such as gRPC, to ensure connectivity between the clients and the server.



\subsection{Monitoring and Visualization} \label{sec-tools-monitoring}

Finally, in this section, monitoring, storage, and data visualization tools are presented.

\subsubsection{Prometheus}\label{sub-tools-prometheus}

Prometheus\footnote{https://prometheus.io/ (accessed April 24th, 2024)} is an open-source time series database, originally developed at SoundCloud, which is used mainly as a monitoring and alerting toolkit. Since its inception in 2012, many companies and organizations have adopted this tool, and the project has a very active developer and user community. It is now a standalone open-source project that is maintained independently of any company. Prometheus was used as the primary application for monitoring the usage of resources in the PoC solution implemented in this study.

\subsubsection{Rook Ceph}\label{sub-tools-rook}
Rook Ceph\footnote{https://rook.io/ (accessed April 24th, 2024)} is a storage solution that combines the capabilities of the Rook storage orchestrator with the Ceph distributed storage system. Rook is an open-source tool for managing storage systems on Kubernetes, while Ceph is a dis\-tribut\-ed object and file storage system that provides scalability, reliability, and performance. Rook Ceph was the persistent storage for the client and server applications.

\subsubsection{Grafana}\label{sub-tools-grafana}

Grafana\footnote{https://grafana.com/ (accessed April 24th, 2024)} is a data visualization tool that is commonly used in conjunction with Prometheus. It allows querying, visualizing, and alerting on metrics collected and stored in a variety of data sources, including Prometheus' time series database. In this study, it was mainly used to create the dashboards to visualize the resource usage metrics of clients and server.

\section{Proof-of-Concept Implementation}\label{chap-poc}

Previous sections detailed the conceptual framework proposed and the tools considered for its implementation. This section details the current PoC implemented to run fully functional FL scenarios in a distributed infrastructure with multiple clients and different data distributions. The source code of the implementations of the software components described in this section, as well as configuration files used to allow the deployment and execution of the experiments discussed in the following sections, are available at our GitHub repository\footnote{
https://github.com/Open-Digital-Twin/framework-fl-testing}.

\begin{figure}[ht]
\centering
\includegraphics[width=\linewidth]{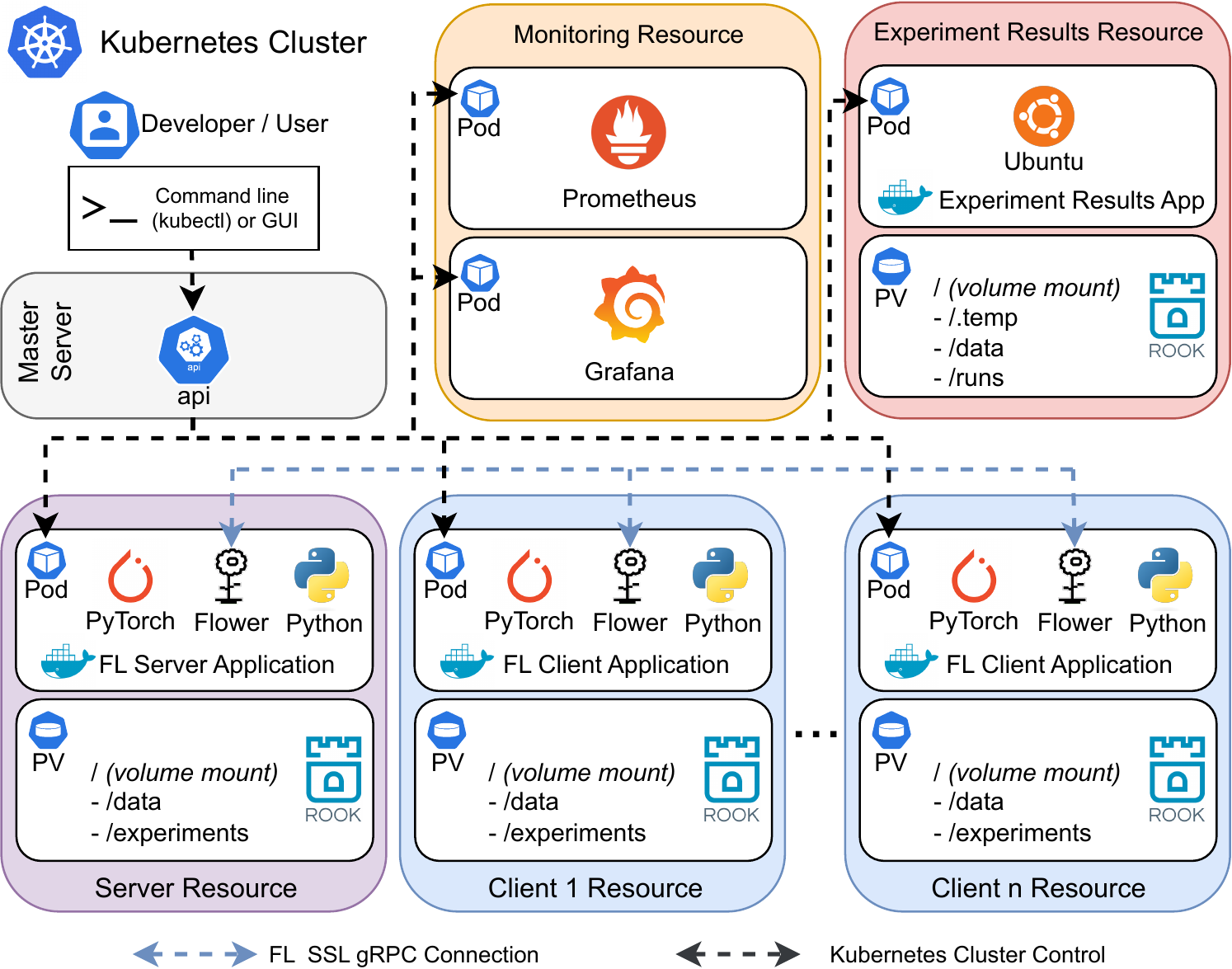}
\caption{Implementation architecture with all tools used in our PoC}
\label{fig:solution-fig}
\end{figure}

The PoC consists of an end-to-end edge-like environment solution using edge computing devices orchestrated by Kubernetes to run the FL applications, the server and the clients, and the applications for monitoring resource usage. Figure \ref{fig:solution-fig} shows the complete diagram of the PoC solution implemented. The distributed infrastructure layer comprises the Kubernetes cluster deployed at the Institute of Informatics, which serves as the foundation for each resource. The application layer can be visualized on top of the resources as containerized Docker images of each application running in Kubernetes pods. Further sections will go through more details about the layers, including how they were assembled with the tools and frameworks presented in Section \ref{chap-tools-framework}, the underlying infrastructure of the Kubernetes cluster, and how the components of the PoC work.

\subsection{Distributed Infrastructure Layer}\label{sec-solution-infrastructure}

In this work, the implementation of the Distributed Infrastructure Layer is primarily based on Kubernetes abstractions. In particular, we used the Kubernetes cluster at the Institute of Informatics, which includes a varied number of computing resources, with 352 CPUs, 544 GiB of RAM, and 6.3 TiB of disk space spread across 43 distinct nodes, as indicated in Table \ref{tab:infrastructure-tab}. These nodes are conveniently categorized into three classifications: ``computer", ``edge" and ``server" enabling experimentation with pods that are constrained to specific machines with hardware that closely matches the real-life devices being simulated. For instance, the ``edge" label encompasses a total of twenty-three Raspberry Pi devices (three Raspberry Pi 4s and twenty Raspberry Pi 3s) with less powerful hardware specifications relative to the ``computer" and ``server" machines. Edge and computer nodes are connected to the campus network at various rooms as any other student lab workstations and laptops, which resembles an actual edge network with fluctuating traffic conditions. On the other hand, server nodes are located in the campus datacenter room, which gives them a more dedicated network connection, especially for traffic among servers.

The cluster offers distributed storage configured using Rook Ceph Filesystem (subsection \ref{sub-tools-rook}) to build a volume abstraction layer on top of the storage resources. This enables the mounting of Persistent Volumes (PV) from the storage pool via a Persistent Volume Claim (PVC) abstractions into the container images that can be shared between applications to enable data distribution. Our server and client applications use these PVCs for saving and sharing training data and experiment results. When a pod requests storage resources, it creates a PVC that specifies the amount of storage required and other requirements, such as access mode and storage class. The Kubernetes scheduler then looks for an available PV that matches the requirements specified in the PVC. If a matching PV is found, it is bound to the PVC, and the pod can use the storage resource provided by the PV. 
The advantage of using PVs and PVCs is that they provide a level of abstraction between the pod and the distributed storage infrastructure, allowing pods to request access and share volumes without having to know the details of the underlying storage infrastructure.

\begin{table*}[ht]
\centering
\begin{tabular}{|l|l|l|l|l|}
\hline
\textbf{Type} & \textbf{Nodes} & \textbf{Total RAM} & \textbf{Total CPUs cores} & \textbf{Total storage space} \\ \hline
Computer      & 14             & 110.9GiB           & 56                        & 587.2GiB  \\ \hline
Edge          & 23             & 40.6GiB            & 92                        & 559.2GiB  \\ \hline
Server        & 6              & 392.4GiB           & 204                       & 5.2TiB    \\ \hline
\end{tabular}
\caption{Institute of Informatics Kubernetes Cluster Capacity}
\label{tab:infrastructure-tab}
\end{table*}



\subsection{Resource Layer}\label{sec-solution-resource}

Kubernetes utilizes labels to organize objects. They are key-value pairs that can be attached to Kubernetes objects such as pods, services, nodes, and deployments and can be used to identify and group related objects. This is a powerful mechanism for selecting and manipulating subsets of objects based on specific criteria and can also be used to manage nodes in a Kubernetes cluster. Nodes are the worker machines that run containerized applications and services in a Kubernetes cluster. By attaching labels to nodes, we can assign specific roles or attributes to them and use those labels to manage and schedule workloads on those nodes. 

For example, one can label nodes based on their hardware characteristics, such as CPU or memory capacity, and then use those labels to schedule workloads that require specific hardware requirements. The Kubernetes Cluster of the Institute of Informatics uses the ``node-type" label to identify if a node is a ``computer", ``server" or ``edge" type of computational device.

Finally, in the PoC solution, the resource layer used is a total reflex of the configuration set in the Kubernetes deployment file. Labels were used to assign where each containerized application runs and how much computing, storage, and networking resources would be available for them to use.

\subsection{Application Layer}\label{sec-solution-application}

The application layer of the PoC solution is composed of all the Docker images built by the author, such as the FL server, client, and experiment results images, which contain all the implemented logic for FL algorithm testing. For monitoring purposes, Prometheus monitoring installed in the Kubernetes cluster was used to capture resource usage by the applications, and Grafana was used to visualize the data in charts. 

\begin{figure*}[ht]
\centering
\includegraphics[width=0.7\linewidth]{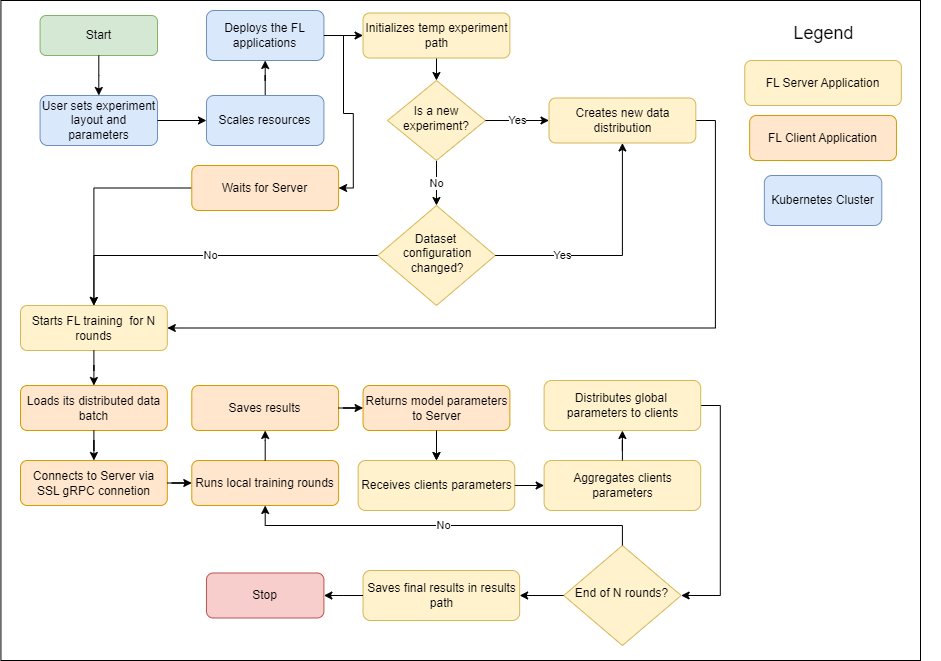}
\caption{High-level flowchart of experiment execution in the PoC solution}
\label{fig:flowchart}
\end{figure*}

Figure \ref{fig:flowchart} shows how an experiment starts from the beginning to the end in the PoC solution from a high-level overview. In the blue boxes, we can see the actions of the Kubernetes cluster, the yellow ones from the FL Server application, and the orange ones from the FL clients. From a high-level perspective, the user configures a scenario -- using command line (i.e., kubectl) or any GUI of their preference -- for deployment, and Kubernetes scales the necessary resources and deploys the applications. Afterward, the FL server handles the data distribution of the dataset and starts the FL training algorithm. The client waits for the server to connect with it, runs local training rounds, and returns the model parameters to the server. The server receives those parameters, aggregates them using the configured aggregation method by the user in the experiment layout, and distributes them back to the clients, who will start local training rounds again with the new parameters. After all server rounds are done, the FL server will save the results in the correct output folder of the experiment, and the experiment will be over.

The GitHub repository used in the development of the solution was organized into isolated packages containing all necessary code, data, or declarations of each component to maintain independence and enable better re-usability. The FL server and client applications were also developed to run in bare-metal environments. In addition, to allow easy application testing during development, you can run a docker-compose\footnote{https://docs.docker.com/compose/ (accessed April 24th, 2024)} environment to deploy the application locally using the built-in docker images. 
Further subsections will detail each one of the applications and their role in the PoC solution.  


\subsubsection{Server Application}\label{subsec-solution-server}

The server is a containerized Python application with the FlowerML server framework as a dependency. The server storage manager is responsible for initializing the experiment root path in the distributed storage, where the results and logs of the currently deployed run of each client will be saved. It will also be responsible for receiving the models trained by the clients, averaging the received parameters using the selected strategy, and then updating the clients' models with the averaged parameters. The connection is done to the clients through a gRPC connection with SSL encryption. The algorithm is outlined in Algorithm 1.

\begin{algorithm}[ht]
 load environment variables with user parameters\;
 \If{$experiment\_path$ don't exists}{
 initialize $experiment\_path$ in storage\;
 create a temporary folder for the current experiment run\;
 initialize experiment configuration file \;
 }
 load experiment configuration file\;
 \If{dataset configuration changed \textbf{or} is empty}{
    load raw data of the dataset\;
    create new data distribution using user parameters\;
    save the data batches in the experiment path \;
    save the new dataset configuration in the configuration file\;
    
 }\Else{
 use current dataset configuration\;
 }
 configure server strategy with the selected $fl\_strategy$\;
 initialize server global model parameters\;
 open connection in $server\_address$\;
 
  \While{$minimum\_clients$ have not yet connected}{
    wait\;
  }
  \For{$i\gets rounds$ \KwTo $0$}{
    receive parameters from clients\;
    aggregate client parameters\;
    send updated parameters back to clients\;
    evaluate current accuracy from clients\;
    }
 save temporary experiment folder with results to current experiment run path\;
 update experiment configuration file\;
 \caption{Server application}
\end{algorithm}

The server address, number of server rounds, dataset, data distribution, global aggregation strategy, client local rounds, and minimum number of connected clients are parameterized for the server application and can be changed in each deployment. 
From the folder structure and source code files available at our GitHub repository, it is easy to correlate each part of the application with the conceptual framework since each source code file encapsulates its corresponding component, as we explain in the following.


\begin{description}
    \item[dataset.py] contains all the implemented classes of the supported datasets by the framework used by the application to handle the dataset configuration in memory and how to obtain the raw data of the dataset properly. At the time of this work, the \textbf{CIFAR-10}, \textbf{CIFAR-100}, and \textbf{FMNIST} datasets were implemented. Further development of this class can enable any dataset to be compatible with the framework. Section~\ref{section-datasets} will detail each one of the currently supported datasets.
    
    \item[main.py] contains the main program loop and the server component of the application, which is responsible for communicating with the clients and controlling the entire FL learning process, which includes selecting available clients to start training; control of the number of FL server rounds and round timeout; global model parameters aggregation and distribution and retrieval of model parameters. All the implemented packages are included here, and the central server program runs as demonstrated in Algorithm 1.
    
    \item[model.py] contains the supported models and is responsible for server-side initialization of global model parameters since some global model aggregation strategies need it. At the time of this work, the current used models for testing are a simple \textbf{CNN}, \textbf{googlenet} \cite{fl-googlenet} and \textbf{resnet} \cite{fl-resnet}. These models were chosen due to the simplicity and popularity. Further development of classes can enable any models to be compatible with the framework.

    \item[storage.py] contains the storage manager class, which is responsible for distributing the raw data of the configured data\-set over the distributed infrastructure storage to achieve user requirements for testing. The storage manager of the server application is also responsible for managing where each experiment will write its data and where the client will output their results data in the system. At the time of this work, the storage manager can create data distributions by changing the following parameters:
    
    \begin{description}
        \item[Balance] boolean parameter specifies if the data should be balanced between clients. If $true$, the dataset is balanced, meaning the data batches have the same number of samples. Otherwise, if $false$, the number of data samples differs between them. 
        \item[Non-IID] boolean parameter specifies if the data should be Non-IID or IID between clients. If $true$, the dataset is Non-IID, meaning there is a class imbalance between the clients. Otherwise, if $false$, the dataset has a balanced class distribution. 
        \item[Distribution] this parameter specifies how data should be distributed regarding the dataset classes if $Non-IID = true$. If set to \textbf{pat}, it will generate a pathological scenario of class imbalance, whereas each client will have only a subset of classes. If set to \textbf{dir:}, it will use a heterogeneous unbalanced Dirichlet distribution of the classes, similar as seen in \cite{fl-dirchlet}, an $\alpha$ parameter can modify that to change the distribution aspects. Further development of this class can enable more data distributions to be compatible with the framework.
    \end{description}
    
    Further development of the storage manager class can enable more types of data distributions to be compatible with the framework.
    
    \item[strategies.py] contains the strategy for supported global model aggregation strategies from the user-specified configuration. At the time of this work, the supported strategies implemented are \textbf{FedAvg}, \textbf{FedAvgM}, \textbf{FedYogi}, and \textbf{FedOpt}.  Further development of this class can enable more strategies to be compatible with the framework.
\end{description}


\subsubsection{Client Application}\label{subsec-solution-client}

The client waits for the server storage manager to initialize the experiment path in the distributed storage, connects to the server through a gRPC connection encrypted with SSL, and performs the pre-defined number of epochs received from the server, which is the number of times that the data set passes through the neural network. When multiple clients are running, an individual client is oblivious to the existence of the other clients -- it can only communicate with the server. The algorithm for the client application can be seen in Algorithm 2.

\begin{algorithm}[ht]
 load environment variables with user parameters
 \While{$experiment\_path$ not initialized by the Server}{
 wait\;
 }
 start the connection with a server in $address$\;
  \While{server connection is open}{
    \For{$i\gets epochs$ \KwTo $0$}{
        iterates through the dataset batch training the local model\;
        saves current results of a round in the experiment path temporary folder\;
    }
    upload model to server\;
    receive updated parameters\;
    update local parameters\;
  }
  
 \caption{Client application}
\end{algorithm}

The client will only upload the model when the server requests the models, and the loop will end when the server finishes its rounds. 
From our folder structure at GitHub, it is easy to correlate each part of the application with the conceptual framework since each source code file encapsulates its corresponding component. The client's models and the address they will connect are parameterized for the client application and can be changed in each deployment.  


\begin{description}
    \item[client.py] contains the client classes with the utilized training and testing algorithms. At the time of this work, it would only support one PyTorch training algorithm.
    
    \item[main.py] contains the client application main loop described in Algorithm 2. It is responsible for initializing the environment variables used to set parameters, such as the local model utilized for training, and common variables, such as the experiment path, to save results.
    
    \item[model.py] contains all the supported models described in the server application section.
    
    \item[storage.py] contains the storage manager class of the client, which is responsible for loading the distributed data batch for training and testing the distributed storage and saving the experiments test results of the client in the storage.
\end{description}


\subsubsection{Monitoring Application}\label{subsec-solution-monitoring}

Prometheus uses the Kubernetes API to discover the various resources it needs to monitor in the cluster, such as pods, services, deployments, nodes, and more. It does this by querying the Kubernetes API server for information about the desired resources and then collecting metrics data from these resources. 

For example, to monitor a Kubernetes pod, Prometheus will query the Kubernetes API server to get information about the pod's name, namespace, labels, and other metadata. Prometheus will then use this information to collect metrics data from the pod, such as CPU and memory usage, network traffic, and other metrics. Grafana queries the data from the Prometheus database to enable real-time visualization of resource usage in dashboards. Some sample charts of these dashboards are presented in Section \ref{section-monitoring}. 


\subsubsection{Experiments Results Application}\label{subsecs-solution-experiment}

\begin{figure}[ht]
\centering
\includegraphics[width=1\linewidth]{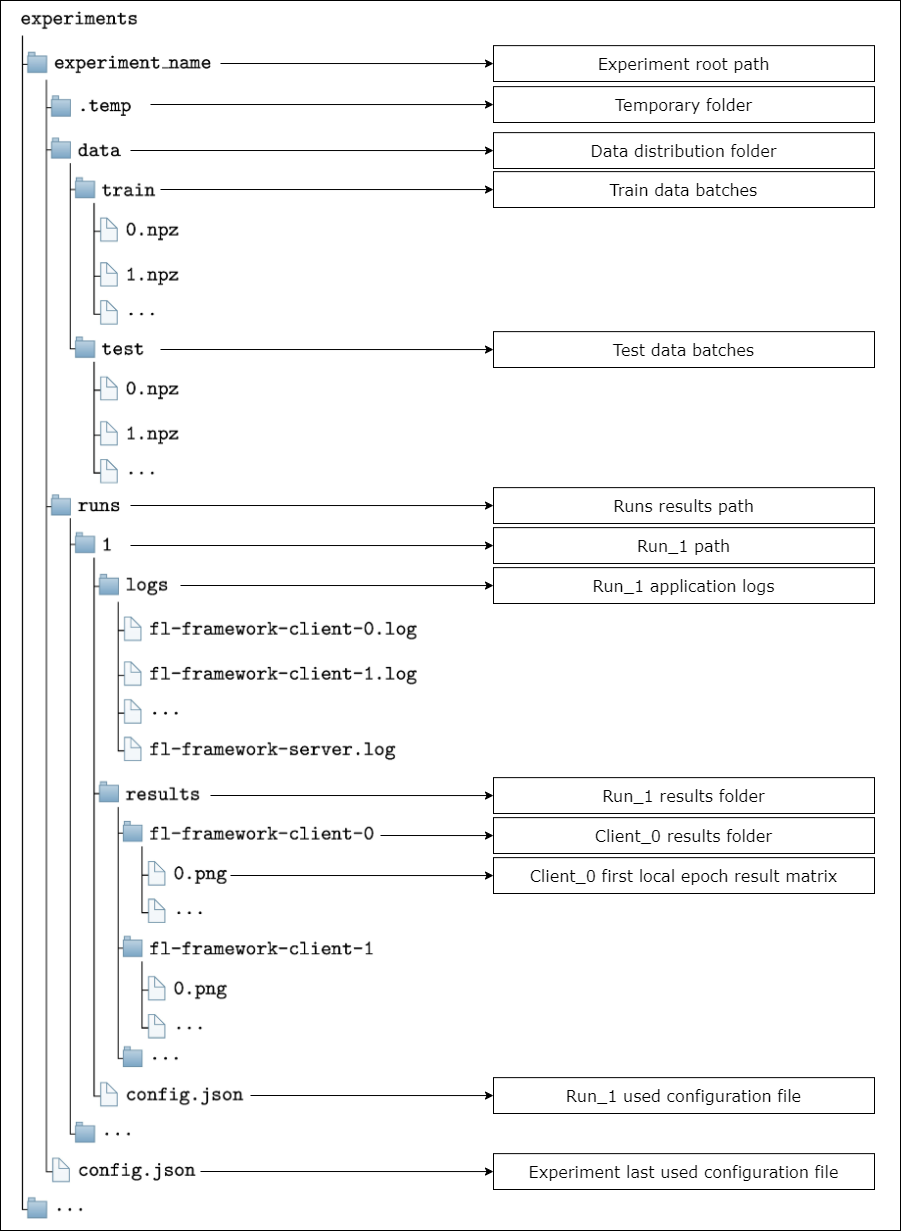}
\caption{Experiments output folder structure}
\label{fig:experiment-directory}
\end{figure}

To retrieve the experiment results from the Kubernetes cluster, a container running an Ubuntu base image from Docker was used to mount the used PVs and access the results. Since the results are stored in files, we can copy them to the local machine to read and interpret them. Figure \ref{fig:experiment-directory} illustrates how the experiments that run in the PoC are saved in the distributed storage.

\begin{description}
    \item[.temp] contains the partial results while running the application. This directory is deleted after the server application storage manager has moved the finalized run into its correct directory. 
    
    \item[data] contains each client's training and testing data batches used in the experiment. 
    
    \item[runs] contains the results for each experiment run. The logs subfolder of a run holds all the logs from the server and client applications, and the results subfolder contains the evaluation matrix for each local epoch run in the clients. 
    A copy of the configuration used in the experiment run is saved to enable visualization of how the data was distributed in each client, how the classes were distributed inside each data batch, and which model and global aggregation strategy was used.
    
    \item[config.json] contains the last configuration used in the experiment. This enables testing of the same data distribution using several parameters, strategies, and models.

\end{description}

\section{Experimental Setup}\label{section-experimental-setup}
This section presents the experimental setup used to evaluate the solution proposed in Section~\ref{chap-solution}. Three experiments were conducted to demonstrate the framework's capabilities. The adjusted parameters are the dataset, global model aggregation, local client model, client epochs, and server rounds. Each experiment was run with ten clients --- named client-0 through client-9 and taken from the set of computational devices labeled \textit{computer} in the cluster --- running with different data distributions. Also, in the Flower framework, the fraction fit was set to 1, and the minimum available clients was set to 10 to ensure running every server round with all the clients participating. The specific clients involved in a given experiment may vary from one run to another, showing how the framework can help reproduce real-life scenarios effectively.

\subsection{Datasets}\label{section-datasets}
The datasets considered for the experiments were the following:
\begin{description}
    \item[CIFAR-10] Consists of 60,000 32x32 color images divided into ten classes, with 6,000 images per class. There are 50,000 training images and 10,000 test images~\cite{cifar}.
    \item[CIFAR-100] Similar to CIFAR-10, except it has 100 classes containing 600 images each. There are 500 training images and 100 testing images per class. The 100 classes in the CIFAR-100 are grouped into 20 superclasses. Each image comes with a \textit{fine} label (the class to which it belongs) and a \textit{coarse} label (the superclass to which it belongs)~\cite{cifar}.
    \item[Fashion-MNIST] A dataset of Zalando's article images consisting of a training set of 60,000 examples and a test set of 10,000 examples. Each example is a 28x28 grayscale image, associated with a label from 10 classes. We intend Fashion-MNIST to be a direct drop-in replacement for the original MNIST dataset~\cite{fl-mnist} for benchmarking machine learning algorithms. It shares the exact image size and structure of training and testing splits.
\end{description}

\subsection{Performance Metrics}\label{section-metrics}

The results from each experiment are saved in performance score matrices and written to log files, as described in Section~\ref{subsecs-solution-experiment}. The F1 score is a standard metric for evaluating classification models and is particularly useful in cases with imbalanced data. It is the harmonic mean of two competing scores: precision and recall. Precision is the proportion of true positives out of all predicted positives (true positives + false positives), measuring how often the model correctly predicts a positive class, and recall is the proportion of true positives out of all actual positives (true positives + false negatives), measuring how well the model can detect positive classes. Thus, for binary classification problems, the F1 score is calculated by

\begin{equation}\label{eq:f1_score}
    \textit{F1 score} = 2 \times \frac{ precision \times recall}{precision + recall} 
\end{equation}

Given the F1 score of each class, as defined by~(\ref{eq:f1_score}), we can aggregate them into different metrics typically used for evaluating the performance of multi-class classification models: the micro, macro, and weighted average F1 scores. These metrics are computed for each client and saved to their log files. In short, the performance metrics analyzed were the following:

\begin{description}
  \item[Accuracy] Measures the model's accuracy as the ratio between the number of correctly predicted samples and the total number of samples in the test dataset. For binary class datasets, the micro average F1 score equals accuracy.
  \item[Macro Averaged F1 Score] Useful when dealing with balanced datasets or evaluating the model's overall performance across all classes equally. It is measured by computing the average F1 score of all classes in the dataset.
  \item [Weighted Averaged F1 Score] Useful for imbalanced datasets, as it assigns more importance to classes with more samples by considering their distribution. It is measured by computing the F1 score of each class and weighting the average F1 score by the number of instances in each class.
  \item[Losses Distributed] Indicates the average loss across all clients. The server's log file contains the average loss of each client.
  \item[Accuracy Distributed] Useful to assess the overall FL accuracy. It is measured by computing the weighted average accuracy of all clients, where the accuracy of each client is weighted by the number of samples in its dataset. The server's log file contains the accuracy of each client.
\end{description}

It should be noted that while prediction performance is a critical metric for any machine learning algorithm, it is not the focus of this work. The primary purpose of the experiments is to demonstrate the framework's capabilities and how it can provide an easy way to completely change an FL testing scenario by only changing input parameters in a deployment file; higher prediction performances would require much more training time, making running numerous experiments in a reasonable time impracticable. It would also require further tweaks on each model run, which is out of the scope of this work.


\section{Results}\label{section-results}

The results were collected directly from the output folder and aggregated into visualization charts to ease analysis. In total, 143 containers (server and clients) were run in the cluster nodes, resulting in 143 log files containing performance measures through each server round. Each experiment run generates 11 log files, one from the server's container and 10 from the clients' containers. As the number of output files can grow exponentially, we considered only the last run of each experiment, i.e., 33 files were analyzed to generate the results.

We used external commercial tools to manipulate the data collected from all log files and show the results through meaningful charts. As a future work, the framework can include automatic chart generation from the performance measures generated from the experiments.  

\subsection{Experiment 1}\label{subsection-experiments-1}

Table~\ref{tab:setup-experiment-1} details the parameters used for the first experiment and the data distribution across clients. The FMNIST dataset was used with a non-IID and unbalanced data distribution. Also, the classes were distributed using a Dirichlet distribution with $\alpha$ = 0.1. The global model aggregation used was FedOpt, and the experiment ran with ten local epochs and ten server rounds.  

\begin{table}[ht]
\centering
\fontsize{5pt}{5pt}
\selectfont
\caption{Configuration parameters and data distribution for Experiment 1}
\label{tab:setup-experiment-1}
\begin{tblr}{
  cell{1-2}{1-5} = {c},
  cell{1}{1} = {c=2}{},
  cell{1}{3} = {c=3}{},
  vline{1-6} = {1-12}{0.03em},
  hline{1,2,3,13} = {1-5}{0.03em},
}
\textbf{Configuration} &  & \textbf{Data Distribution} &  &  \\
\textbf{Parameter} & \textbf{Value} & \textbf{Client} & \textbf{Train Batch Size} & \textbf{Test Batch Size}  \\
Dataset            & FMNIST            & client-0 & 6345 & 2116 \\
Experiment Name    & kubernetes-test-1 & client-1 & 135  & 46    \\
FL Strategy        & FedOpt            & client-2 & 4137 & 1380 \\
Model              & Multilayer Perceptron (MLP)              & client-3 & 600  & 200   \\
Data Balance       & False             & client-4 & 6109 & 2037 \\
Non-IID            & True              & client-5 & 7389 & 2464 \\
Class Distribution & Dirchlet          & client-6 & 8968 & 2990 \\
$\alpha$           & 0.1               & client-7 & 5734 & 1912 \\
Server Rounds      & 10                & client-8 & 6665 & 2222 \\
Local Rounds       & 10                & client-9 & 6413 & 2138 \\
\end{tblr}
\end{table}

\begin{figure*}
    \centering
    \begin{subfigure}[b]{0.45\linewidth}
        \centering
        \includegraphics[width=\linewidth]{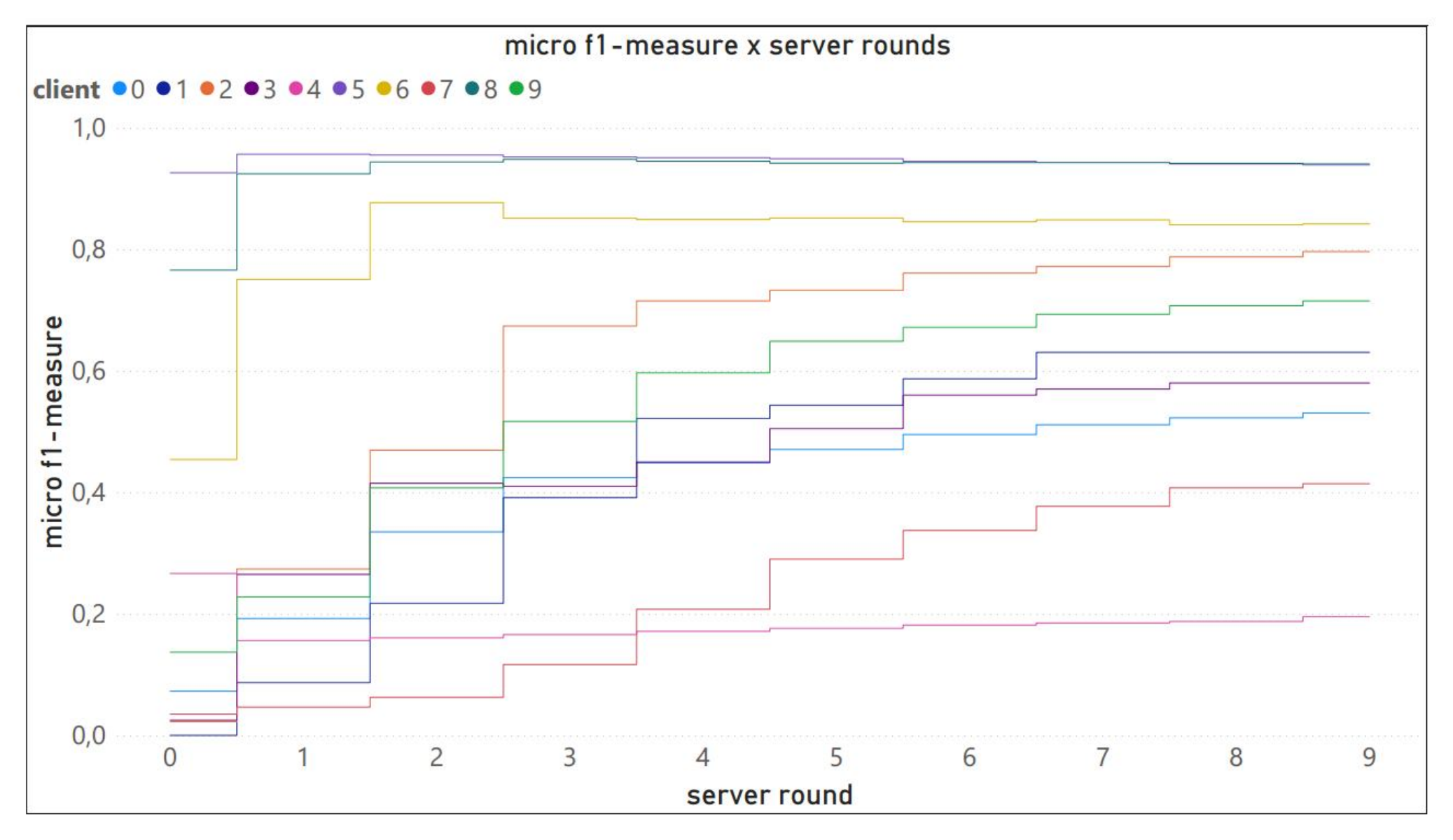}
        \caption{Evolution of each client's Micro F1 score throughout the server rounds.}
        \label{fig:experiment_1_micro}
    \end{subfigure}
    \begin{subfigure}[b]{0.45\linewidth}
        \centering
        \includegraphics[width=\linewidth]{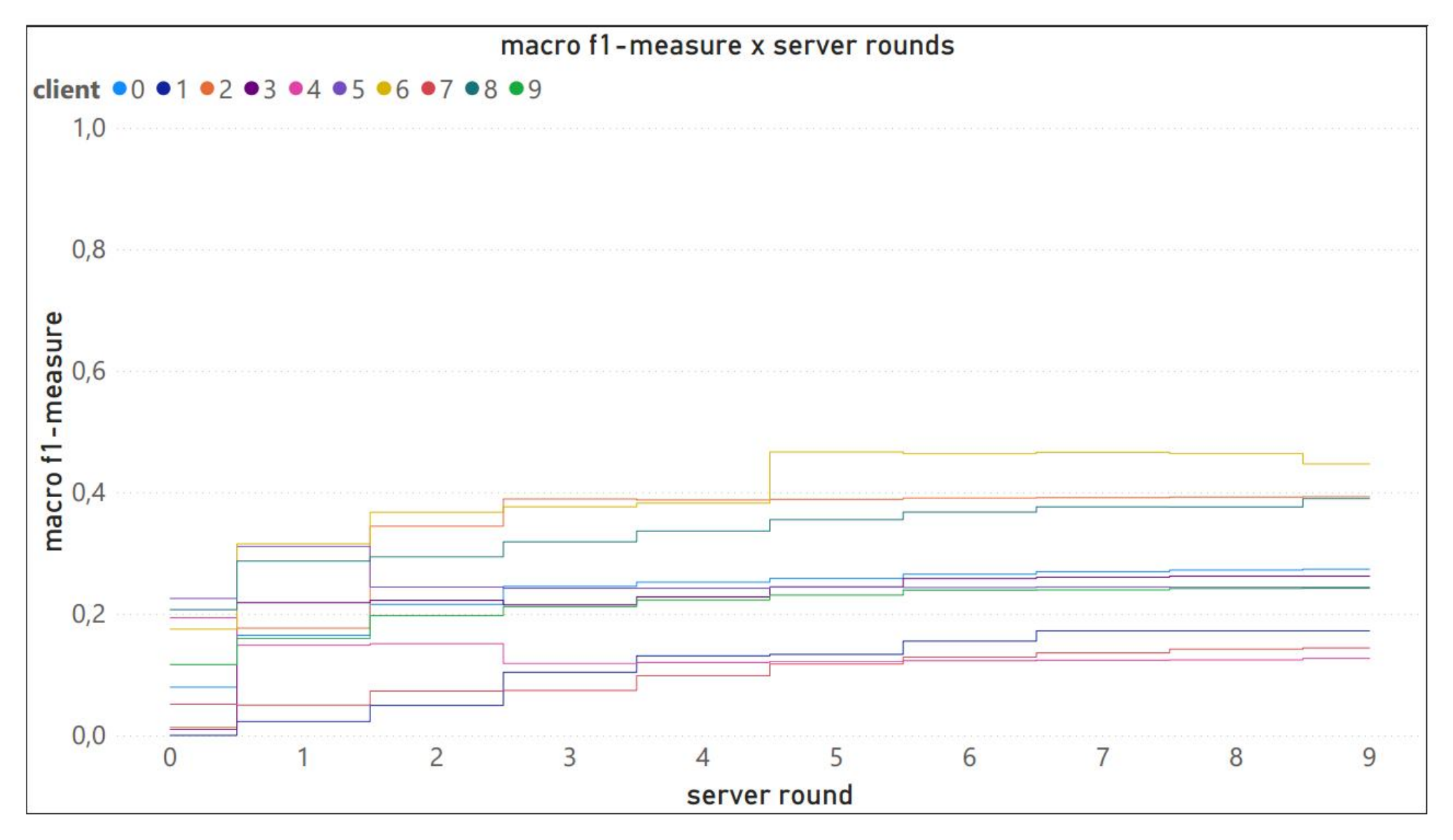}
        \caption{Evolution of each client's Macro F1 score throughout the server rounds.}
        \label{fig:experiment_1_macro}
    \end{subfigure}
    \vskip\baselineskip
    \begin{subfigure}[b]{0.45\linewidth}
        \centering
        \includegraphics[width=\linewidth]{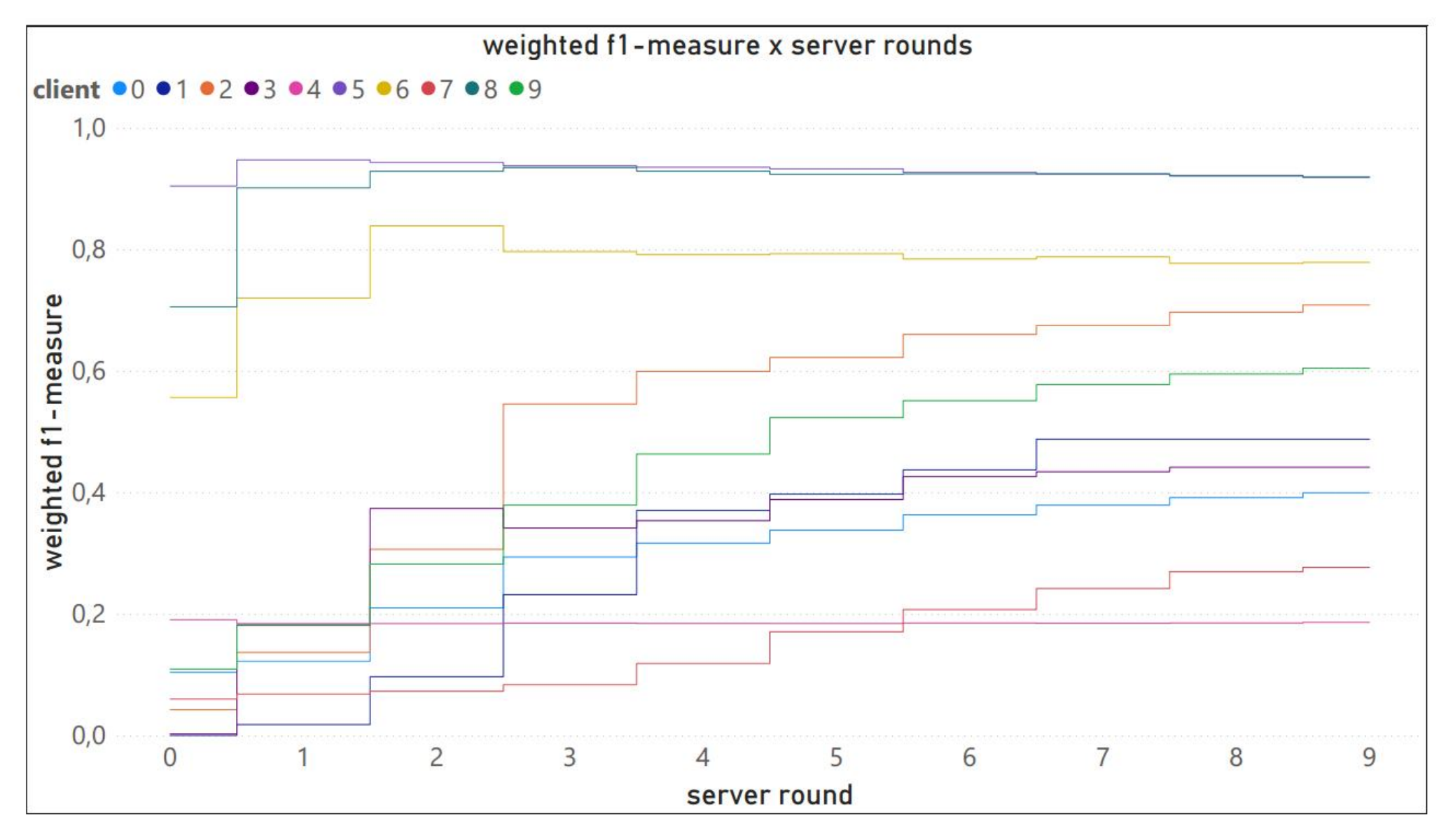}
        \caption{Evolution of each client's Weighted F1 score throughout the server rounds.}
        \label{fig:experiment_1_weighted}
    \end{subfigure}
    \begin{subfigure}[b]{0.45\linewidth}
        \centering
        \includegraphics[width=\linewidth]{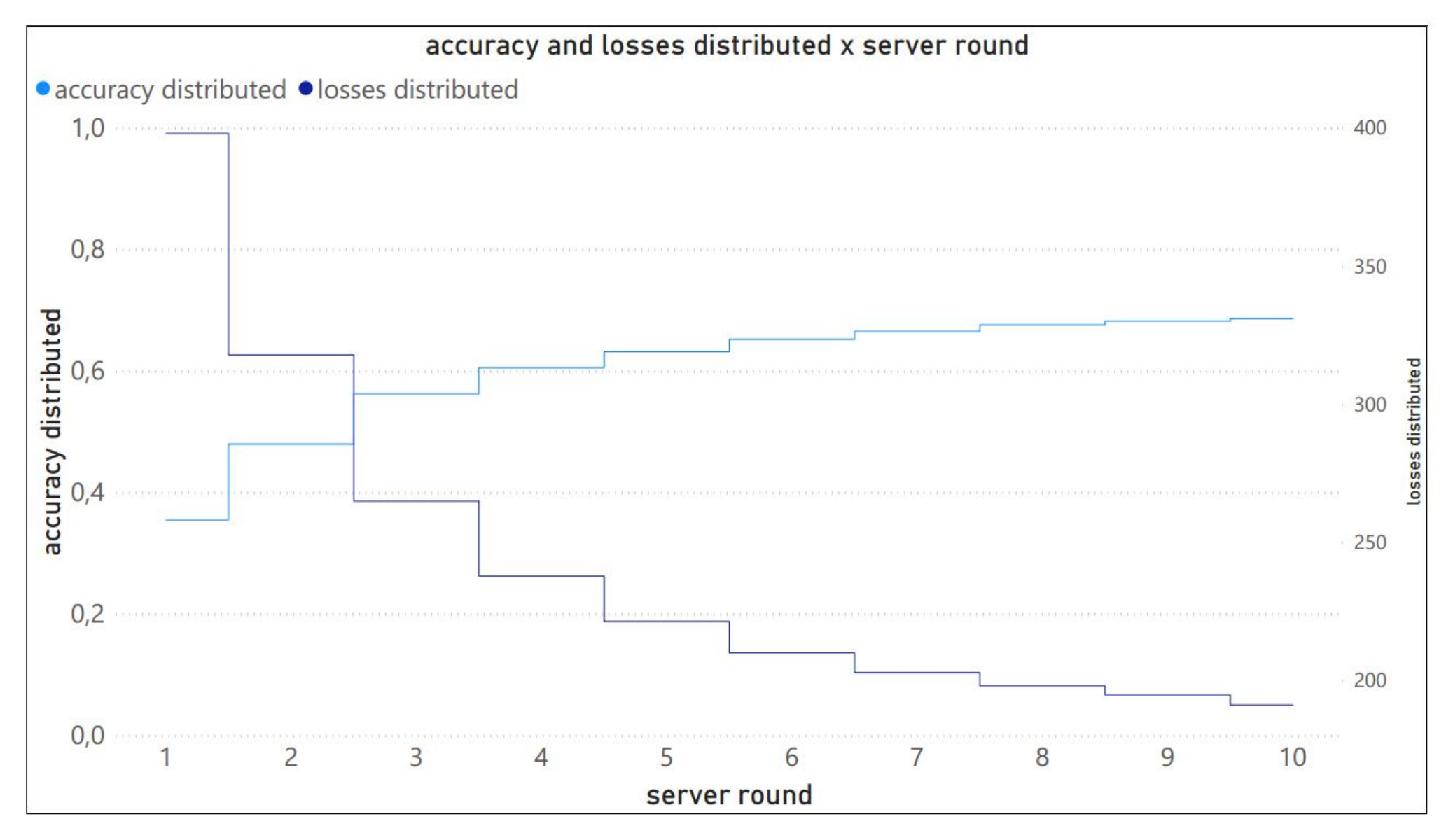}
        \caption{Accuracy and losses distributed throughout the server rounds.}
        \label{fig:experiment_1_server}
    \end{subfigure}
    \caption{Results from Experiment 1}
    \label{results:experiment-1}
\end{figure*}

Figure~\ref{fig:experiment_1_micro} shows each client's \emph{micro F1 score} through the server rounds. The x-axis indicates the server round number, and the y-axis indicates the computed micro f1 score. It is important to notice that micro f1 score measures the percentage of correctly classified cases regardless of their classes, assigning the same importance to all instances. The figures indicate that clients with large amounts of data had a relatively high initial performance in the first rounds. In contrast, the performance of clients with little data was initially low. Over the server rounds, the chart reveals that the aggregate model had a subtly negative impact on the performance of clients with more data available while benefiting the performance of clients with less data. 

Figure~\ref{fig:experiment_1_macro} shows the macro F1 score of each client through the server rounds. Notice that macro F1 assigns the same importance to each class of the dataset, which means this metric tends to exhibit better results as the performance of all classes in the dataset improves evenly. The results suggest that the performance may seem unacceptable when focusing on averaging the performance of different classes with the same weight, despite some clients having a reasonable micro f1 score, as shown in Figure~\ref{fig:experiment_1_micro}. Even so, as a general trend, the chart reveals performance increases over the server rounds.  

Another way to evaluate the model's performance is to measure the clients' weighted F1 score, i.e., the average F1 score weighted by the number of instances in each class, as exemplified in Figure~\ref{fig:experiment_1_weighted}. The chart shows that some clients (5, 6, and 8) had a relatively high initial performance with a slight decrease over time. Contrasting with the results in~Figure \ref{fig:experiment_1_macro}, we can note that the performance of these clients in the majority classes (with more instances) is considerably higher in the initial rounds. In these majority classes, however, the aggregated model worsened the performance over the server runs, even though the overall performance considering all classes improved. When observing client-6 in Figure~\ref{fig:experiment_1_macro}, for example, there is a performance leap between rounds 4 and 5, suggesting that the aggregated model improved the model's overall performance despite reducing performance for the majority class.

Figure~\ref{fig:experiment_1_server} shows the distributed accuracy over all clients and the distributed loss of the FedOpt aggregation strategy in each server round. The x-axis indicates the server round number, the left y-axis indicates the accuracy, and the second y-axis indicates the losses. We can see that the overall performance of the FL algorithm converged to an accuracy between 60\% and 70\% with less than 100 of loss. 

Further analysis could be done to understand better how the dataset's class distribution impacted the results, as the framework saves this information in the configuration file of each experiment run. Recall that the purpose of this work is not to present a comprehensive evaluation of FL algorithms but to demonstrate how the proposed framework can aid in doing so. Therefore, further analysis can be subject of future work.  

\subsection{Experiment 2}\label{subsection-experiments-2}
As shown in Table~\ref{tab:setup-experiment-2}, the second experiment was run using the CIFAR-10 dataset with non-IID and unbalanced data distribution, similar to the first experiment. However, we used a pathological class distribution for two classes per client and the FedAvg global model aggregation. The experiment ran with fifteen local epochs and ten server rounds. The results analysis follows the same approach used in Section~\ref{subsection-experiments-1}.

\begin{figure*}[ht!]
    \centering
    \begin{subfigure}[b]{0.45\linewidth}
        \centering
        \includegraphics[width=1\linewidth]{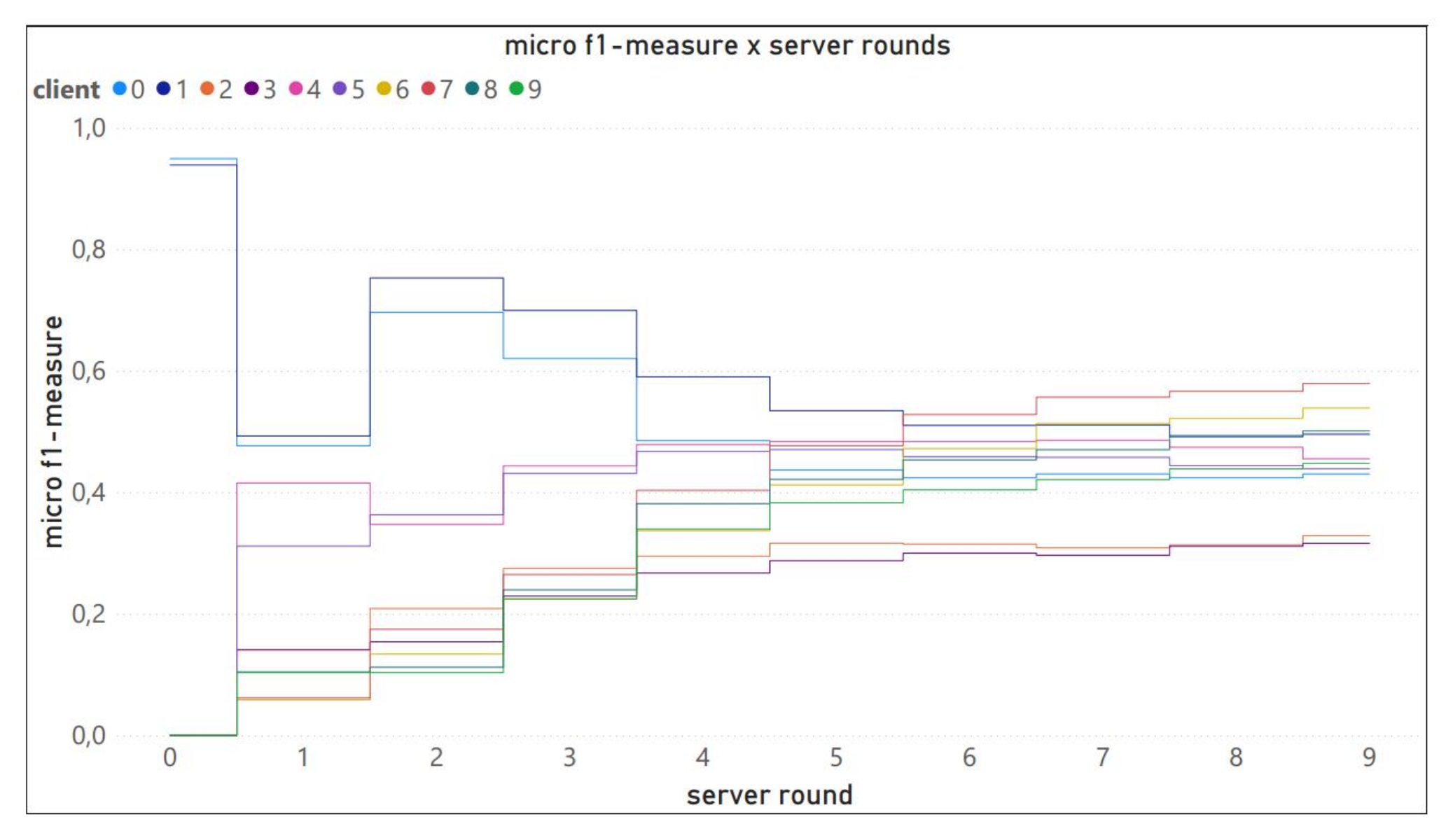}
        \caption{Evolution of each client's Micro F1 score throughout the server rounds.}
        \label{fig:experiment_2_micro}
    \end{subfigure}
    \begin{subfigure}[b]{0.45\linewidth}
        \centering
        \includegraphics[width=1\linewidth]{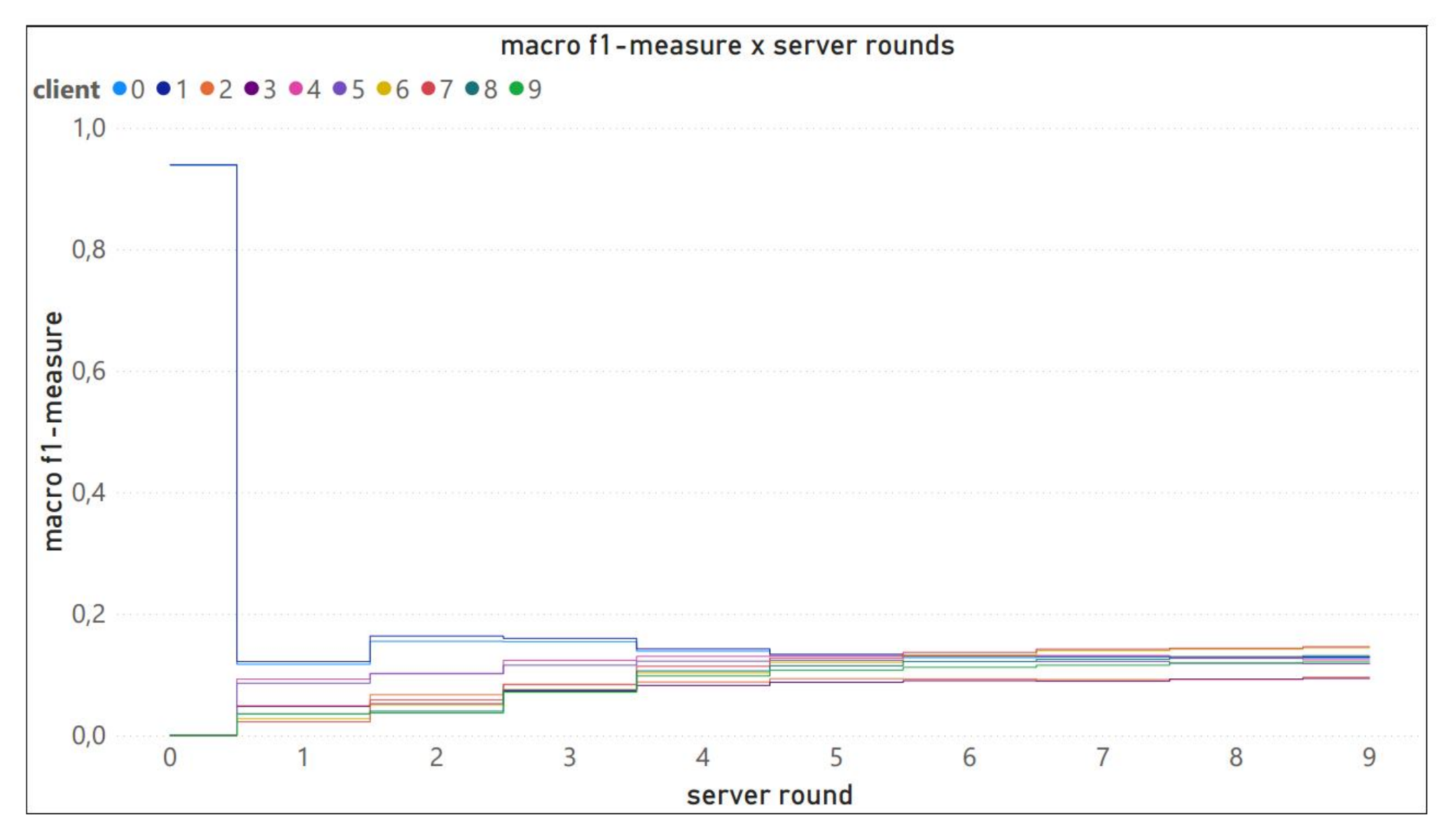}
        \caption{Evolution of each client's Macro F1 score throughout the server rounds considering all classes.}
        \label{fig:experiment_2_macro}
    \end{subfigure}
    \vskip\baselineskip
    \begin{subfigure}[b]{0.45\linewidth}
        \centering
        \includegraphics[width=1\linewidth]{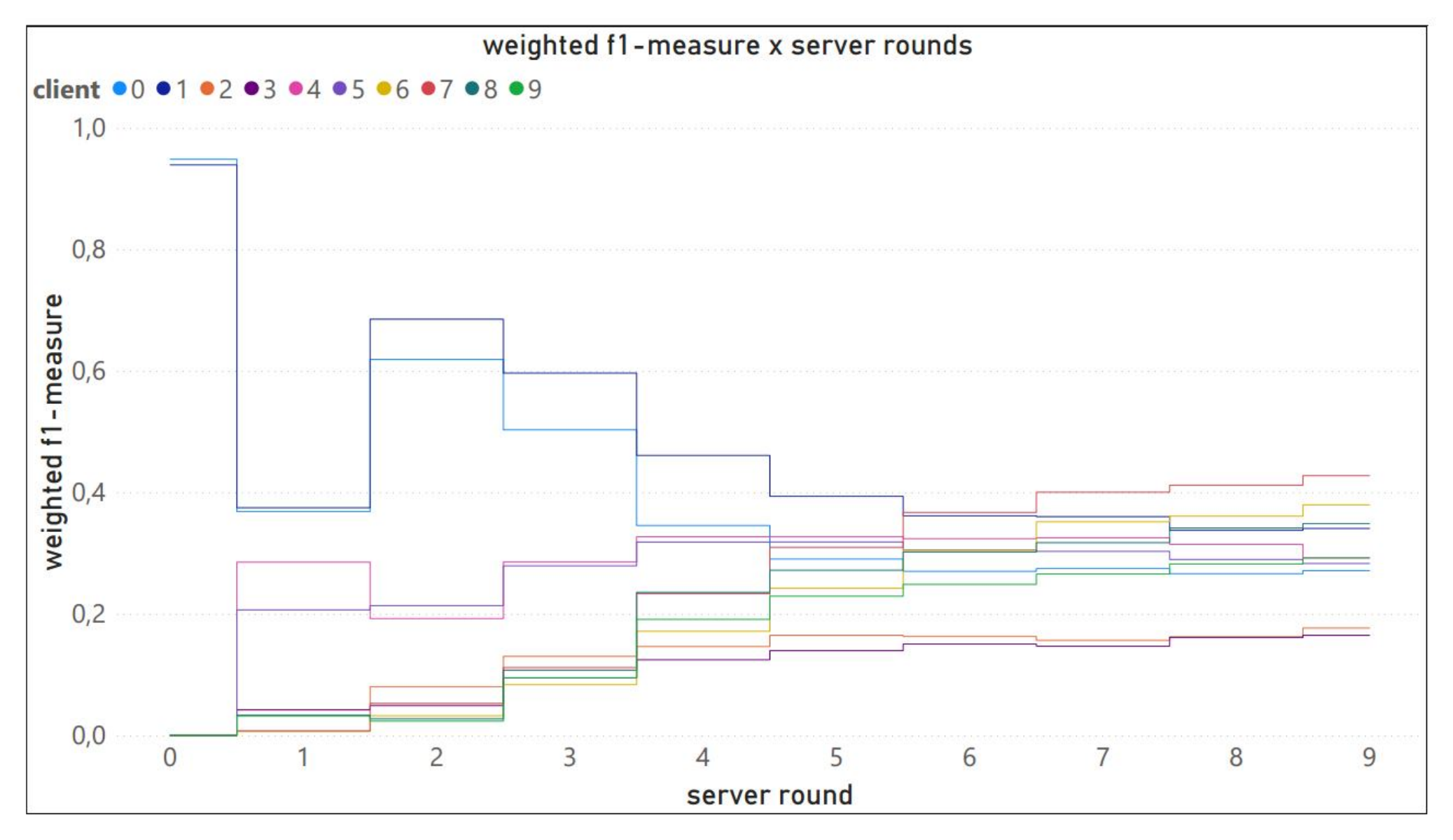}
        \caption{Evolution of each client's Weighted F1 score throughout the server rounds.}
        \label{fig:experiment_2_weighted}
    \end{subfigure}
    \begin{subfigure}[b]{0.45\linewidth}
        \centering
        \includegraphics[width=1\linewidth]{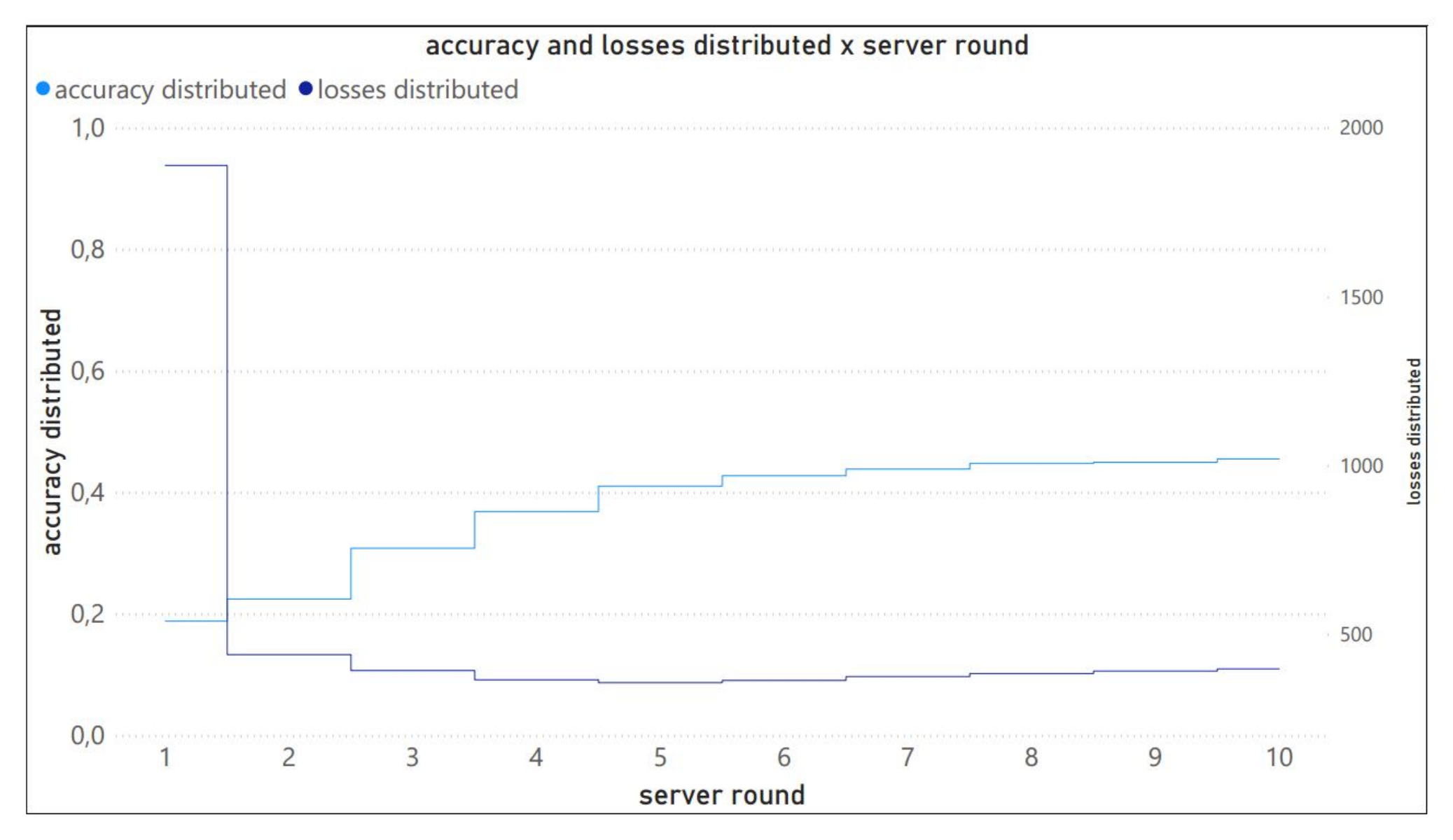}
        \caption{Accuracy and losses distributed throughout the server rounds.}
        \label{fig:experiment_2_server}
    \end{subfigure}
    \caption{Results from Experiment 2}
    \label{results:experiment-2}
\end{figure*}

\begin{table}[ht]
\centering
\fontsize{6pt}{6pt}
\selectfont
\caption{Configuration parameters and data distribution for Experiment 2}
\label{tab:setup-experiment-2}
\begin{tblr}{
  cell{1-2}{1-5} = {c},
  cell{1}{1} = {c=2}{},
  cell{1}{3} = {c=3}{},
  vline{1-6} = {1-12}{0.03em},
  hline{1,2,3,13} = {1-5}{0.03em},
}
\textbf{Configuration} &  & \textbf{Data Distribution} &  &  \\
\textbf{Parameter} & \textbf{Value} & \textbf{Client} & \textbf{Train Batch Size} & \textbf{Test Batch Size}  \\
Dataset            & CIFAR-10          & client-0 & 1422 & 474  \\
Experiment Name    & kubernetes-test-2 & client-1 & 7578 & 2526 \\
FL Strategy        & FedAvg            & client-2 & 1950 & 651  \\
Model              & CNN               & client-3 & 7049 & 2350 \\
Data Balance       & False             & client-4 & 1278 & 426  \\
Non-IID            & True              & client-5 & 7722 & 2574 \\
Class Distribution & Pathological      & client-6 & 2463 & 822  \\
Class per Client   & 2                 & client-7 & 6536 & 2179 \\
Server Rounds      & 10                & client-8 & 2326 & 776  \\
Local Rounds       & 15                & client-9 & 6673 & 2225 \\
\end{tblr}
\end{table}

Figure~\ref{fig:experiment_2_micro} shows the micro f1 score of all clients through the server rounds. We can see that clients 0 and 1 can initially classify a good part of the test instances, but most clients cannot classify any cases. As more server rounds are run, however, the performance of clients with high micro f1 score decreases while the performance of clients with low micro f1 score increases. As a result, the overall performance stabilizes in a medium range between 30\% and 60\%, i.e., the aggregation model was able to distribute the performance among clients.

Figure~\ref{fig:experiment_2_macro} presents the macro F1 score of all clients during the server rounds. It shows that clients 0 and 8 initially managed to obtain an excellent performance in the two classes from which they had data. However, aggregation severely affected their performance, reducing it to less than 20\%. However, aggregation positively affected clients who performed quite poorly initially. Nevertheless, the general performance of all clients, considering all their classes equally important, turned out to be very low.

In contrast, Figure~\ref{fig:experiment_2_weighted} shows the weighted F1 score, suggesting more erratic behavior during server rounds. We can note that clients 0 and 8, which had a notable decrease in performance in the first round when all classes had the same weight, as shown in Figure~\ref{fig:experiment_2_macro}, had a different behavior when considering the importance of each class proportional to the number of instances. The performance of these clients for their majority classes dropped after round 0 to less than 40\% but increased after round 1 to between 60\% and 70\% and then dropped again throughout the server rounds. Deeper analyses must be performed to investigate this behavior. Conversely, we also note that the performance of clients who had poor performance at the beginning had consistent performance improvements over the server rounds. In general, the average performance of all clients stabilized in the range between 30\% and 40\%.

Figure~\ref{fig:experiment_2_server} shows the distributed accuracy over all clients and the aggregated loss of the FedAvg strategy in each server round. The overall performance of the FL algorithm converged to an accuracy between 40\% and 50\% with almost 300 of loss.

\begin{figure*}[ht!]
    \centering
    \begin{subfigure}[b]{0.45\linewidth}
        \centering
        \includegraphics[width=1.0\linewidth]{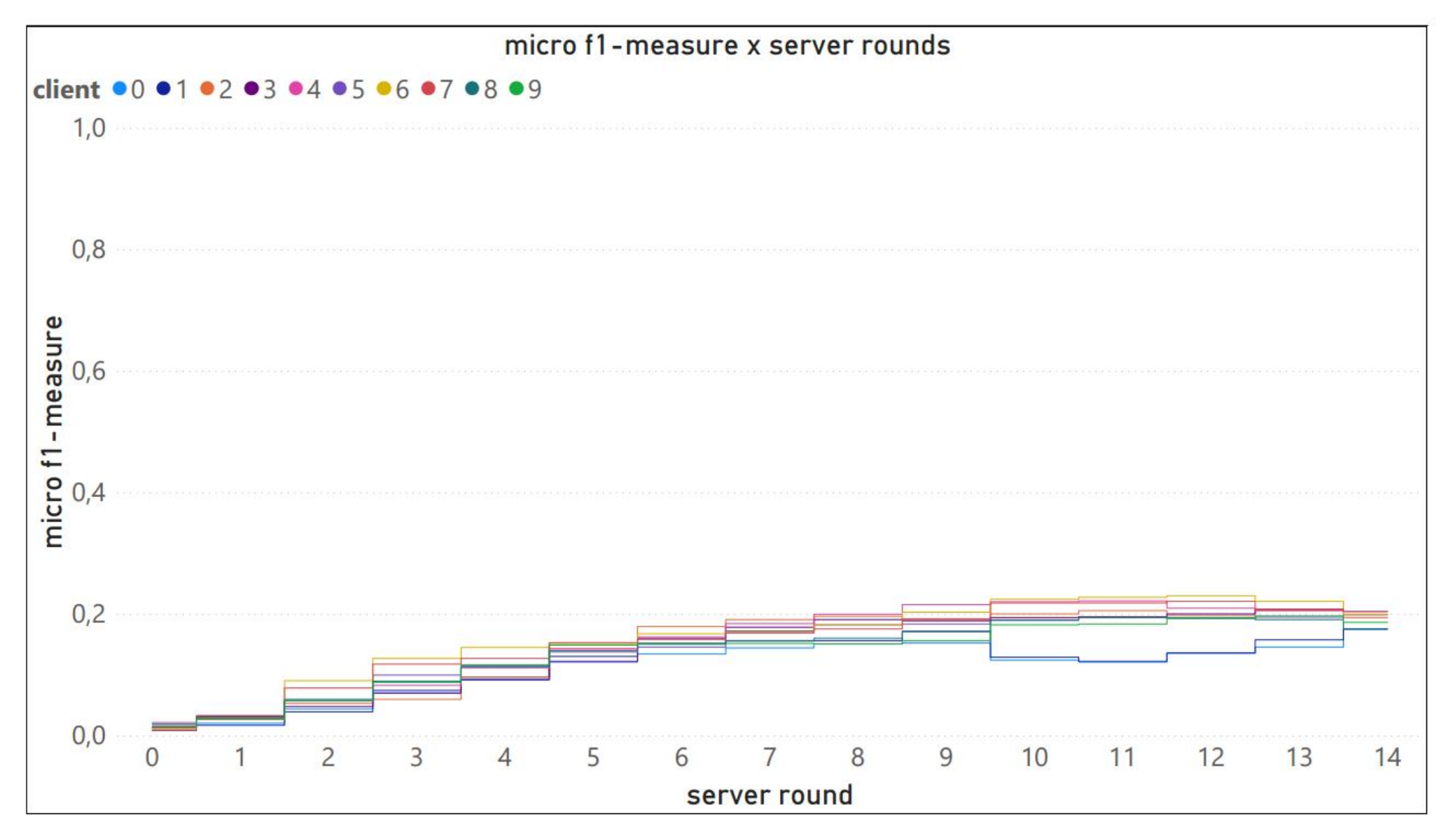}
        \caption{Evolution of each client's Micro F1 score throughout the server rounds.}
        \label{fig:experiment_3_micro}
    \end{subfigure}
    \begin{subfigure}[b]{0.45\linewidth}
        \centering
        \includegraphics[width=1.0\linewidth]{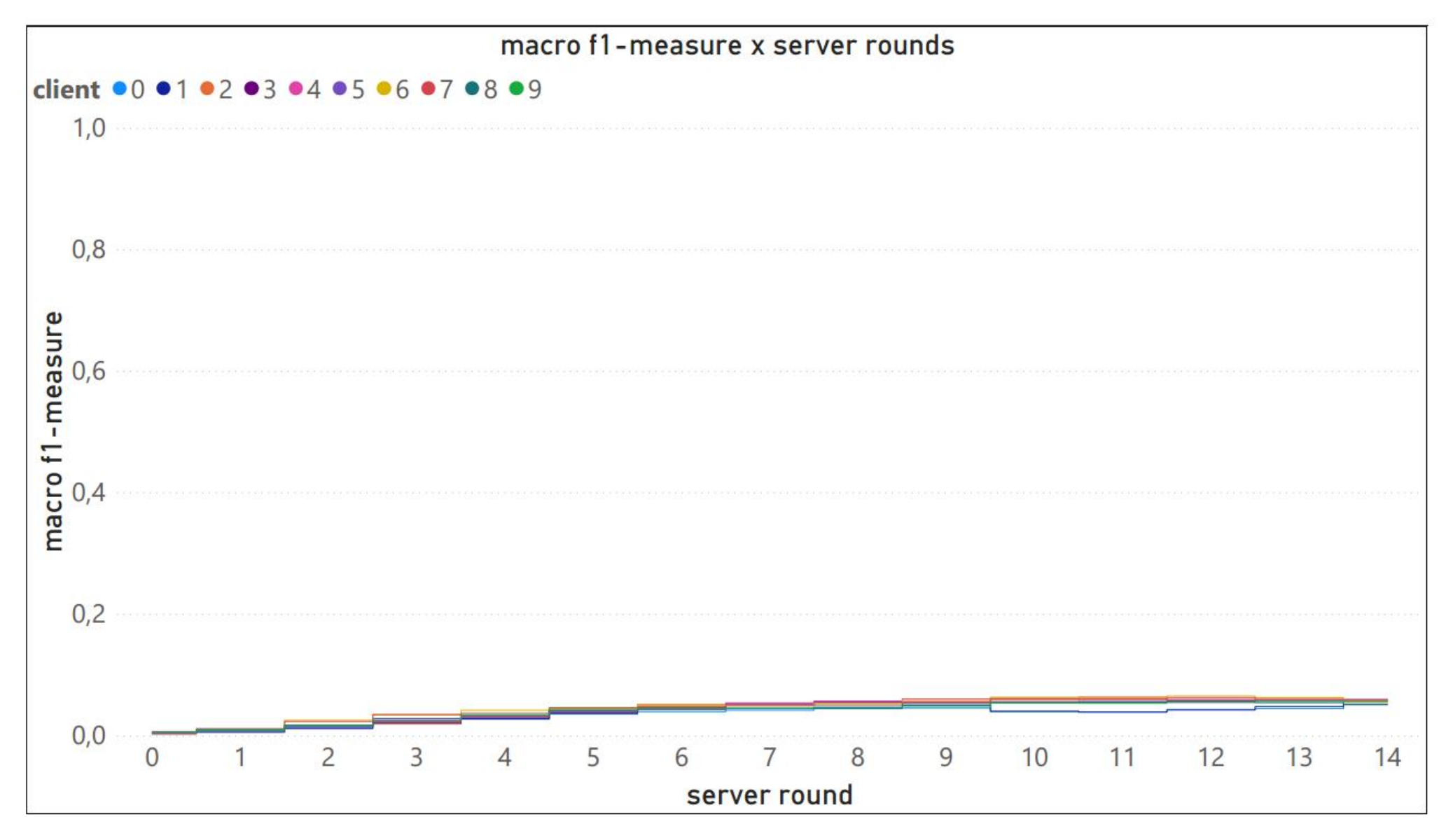}
        \caption{Evolution of each client's Macro F1 score throughout the server rounds.}
        \label{fig:experiment_3_macro}
    \end{subfigure}
    \vskip\baselineskip
    \begin{subfigure}[b]{0.45\linewidth}
        \centering
        \includegraphics[width=1.0\linewidth]{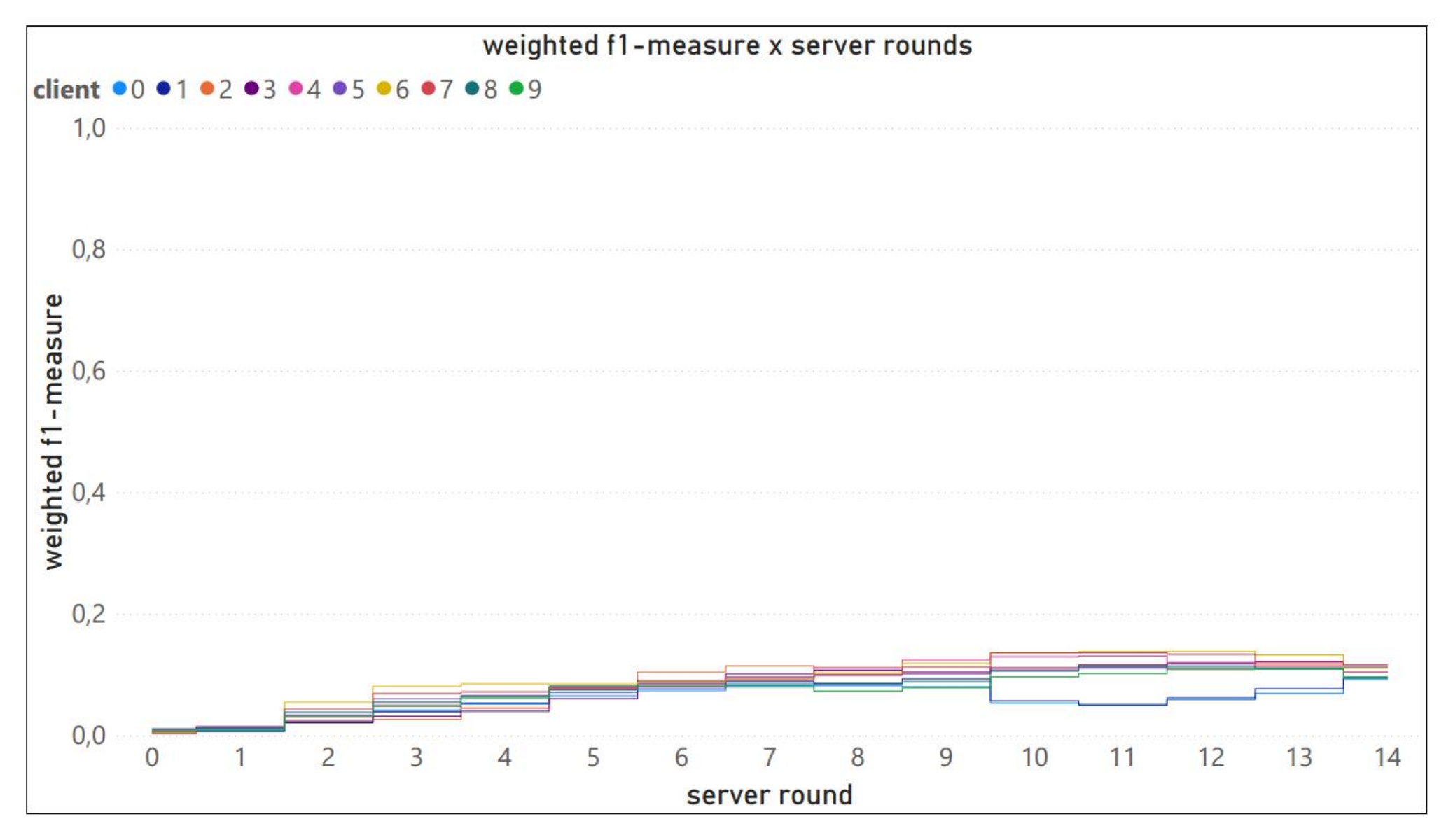}
        \caption{Evolution of each client's Weighted F1 score throughout the server rounds.}
        \label{fig:experiment_3_weighted}
    \end{subfigure}
    \begin{subfigure}[b]{0.45\linewidth}
        \centering
        \includegraphics[width=1.0\linewidth]{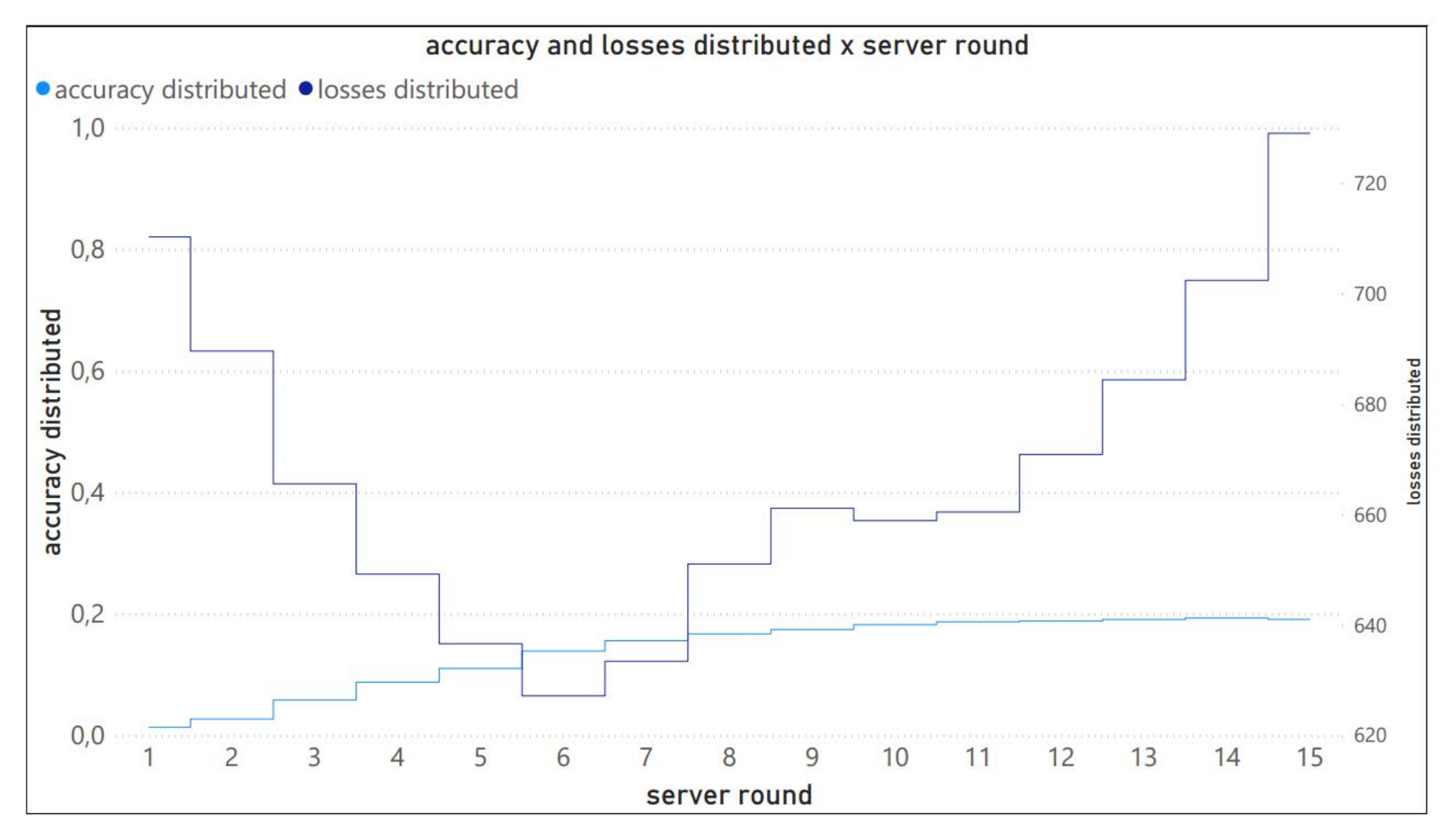}
        \caption{Accuracy and losses distributed throughout the server rounds.}
        \label{fig:experiment_3_server}
    \end{subfigure}
    \caption{Results from Experiment 3}
    \label{results:experiment-3}
\end{figure*}

\subsection{Experiment 3}\label{subsection-experiments-3}

The last experiment was run using the CIFAR-100 dataset, which also had non-IID but a balanced data distribution. Similar to the second experiment, a pathological distribution of twenty classes per client was used. The global model aggregation used was FedYogi, and the experiment ran with ten local epochs and fifteen server rounds. Table \ref{tab:setup-experiment-3} shows the parameters and the balanced data distribution used in the experiment.

\begin{table}[ht]
\centering
\fontsize{6pt}{6pt}
\selectfont
\caption{Configuration parameters and data distribution for Experiment 3}
\label{tab:setup-experiment-3}
\begin{tblr}{
  cell{1-2}{1-5} = {c},
  cell{1}{1} = {c=2}{},
  cell{1}{3} = {c=3}{},
  vline{1-6} = {1-12}{0.03em},
  hline{1,2,3,13} = {1-5}{0.03em},
}
\textbf{Configuration} &  & \textbf{Data Distribution} &  &  \\
\textbf{Parameter} & \textbf{Value} & \textbf{Client} & \textbf{Train Batch Size} & \textbf{Test Batch Size}  \\
Dataset            & CIFAR-100         & client-0 & 4500 & 1500 \\
Experiment Name    & kubernetes-test-3 & client-1 & 4500 & 1500 \\
FL Strategy        & FedYogi           & client-2 & 4500 & 1500 \\
Model              & resnet            & client-3 & 4500 & 1500 \\
Data Balance       & True              & client-4 & 4500 & 1500 \\
Non-IID            & True              & client-5 & 4500 & 1500 \\
Class Distribution & Pathological      & client-6 & 4500 & 1500 \\
Class per Client   & 20                & client-7 & 4500 & 1500 \\
Server Rounds      & 15                & client-8 & 4500 & 1500 \\
Local Rounds       & 10                & client-9 & 4500 & 1500  
\end{tblr}
\end{table}

The experiment deals with a classification problem that is fairly difficult in itself because there are more classes to predict, and originally, there were relatively few instances per class available. Furthermore, in the distributed context, each client has access to only 20 of the classes and less than half of the original dataset samples per class. Due to this, we expect poor performance in this scenario.

The chart in Figure \ref{fig:experiment_3_micro} shows that the proportion of predicted instances generally increases for all clients, with some exceptions (see between rounds 9 and 13), demonstrating that, in general, aggregation benefits clients over time. 

Figure \ref{fig:experiment_3_macro} shows the f1-macro measure. Thus, considering the performance for all classes and those with equal importance, the general performance of all clients is poor, not even reaching 10\% of the macro f-measure. However, it is possible to verify the overall beneficial effects of aggregation across the rounds, albeit with some exceptions.

Figure \ref{fig:experiment_3_weighted}, with the f1-weighted measure, shows a scenario similar to that seen in charts 6.9 and 6.10. That is, in general, the performance of most clients improves with aggregation over the rounds, with some exceptions. However, in this scenario, the classes represented in each client gain more importance, which is responsible for increasing the performance in this metric compared to the macro average.

Figure \ref{fig:experiment_3_server} shows the aggregated accuracy over all the clients and the aggregated loss of the strategy in each server round. The X-axis indicates the server round number, the left Y-axis indicates the accuracy, and the second Y-axis shows the losses. It can be noted that the overall performance of the FL algorithm converged between 40-45\% accuracy and closer to 800 of loss. It is curious how the loss of the strategy decreased in the beginning and afterward increased to a higher level than it began. This may signal the model's difficulty in maintaining consistent improvements when aggregating updates from clients with varying data distributions. After a certain point (server round 6), the model may be overfitting to specific client data, leading to increased loss on more diverse or unseen data. Additionally, in this FL scenario, models from different clients with heterogeneous (non-IID) data distributions are aggregated. While the model may perform well on certain clients initially, it may struggle to generalize as updates from clients with different data distributions are integrated, which can contribute to the rise in loss. This behavior suggests a potential need to explore alternative aggregation strategies or methods that handle non-IID data more effectively. 

\section{Resource Usage Monitoring}\label{section-monitoring}

The PoC solution uses the Grafana application to monitor the use of computing and networking resources during the experiments, allowing the visualization of the data collected by the Prometheus. To demonstrate some of the capabilities of the PoC solution on monitoring, an analysis of CPU usage and network traffic was made in Experiment 3 (subsection \ref{subsection-experiments-3}). It is important, once again, to emphasize the focus on the demonstration of the capabilities and not on the achieved performance. Similar analyzes can be done for all the experiments run since all the metrics are saved in the Prometheus server database.

\begin{figure*}[ht!]
\centering
\includegraphics[width=0.7\linewidth]{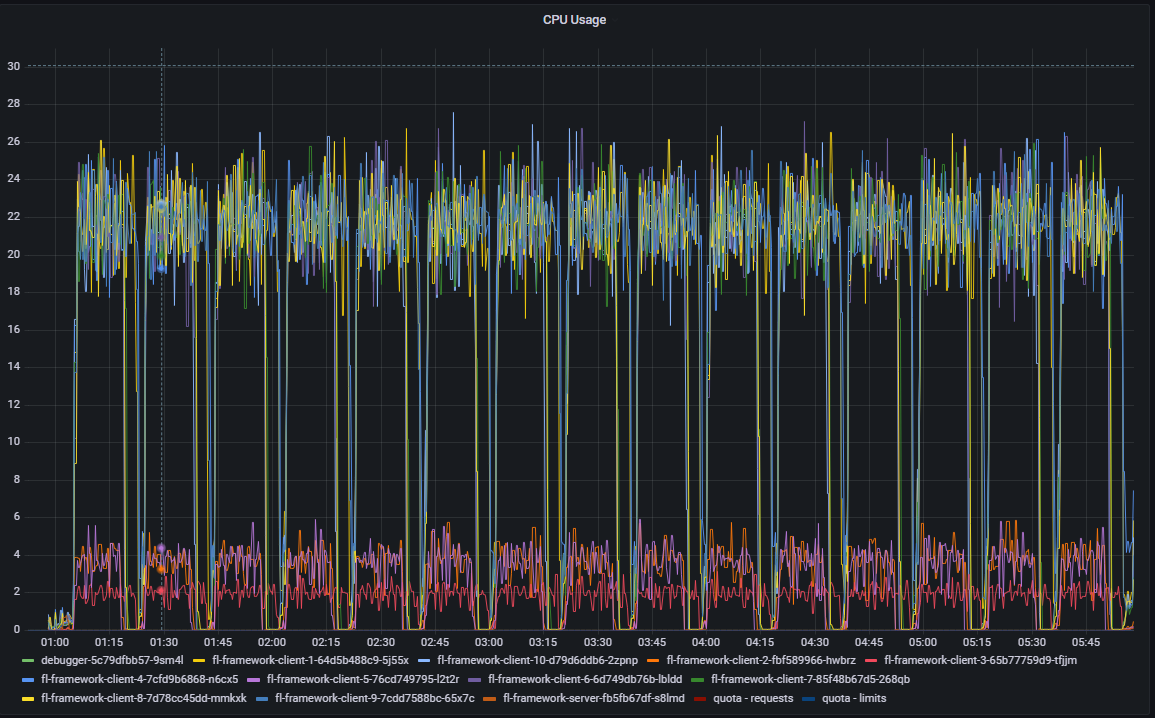}
\caption{CPU Usage from server and clients in experiment 3}
\label{fig:cpu_all_3}
\end{figure*}

\subsection{CPU Usage}\label{subsection-CPU-usage}

Figure \ref{fig:cpu_all_3} shows each client and server's CPU resource usage in the third experiment. The X-axis indicates the timestamp of the metric, and the Y-axis indicates the CPU usage in percentage, and each data sample has a 10-second interval between one another.

\begin{figure*}[ht!]
\centering
\includegraphics[width=0.7\linewidth]{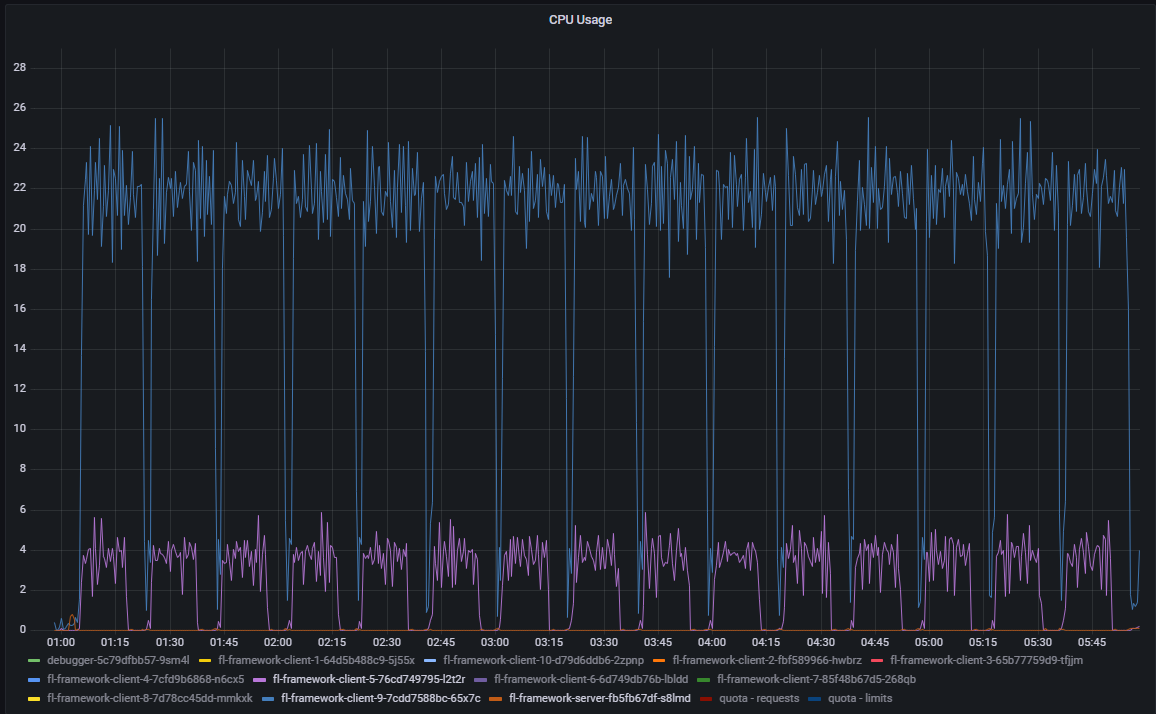}
\caption{CPU Usage from two clients with different hardware in experiment 3}
\label{fig:cpu_different_3}
\end{figure*}

We can notice that the figure is quite chaotic since all the ten clients and the server are plotted in it. However, we can see the execution of each of the fifteen server rounds ran in the experiment and that some of the clients had a CPU usage of 20-30\% while others stayed between 1-5\%. This can be explained by the client's CPU processing power heterogeneity since some clients run on different hardware profiles. We also noticed that the workload was distributed evenly across the clients since the data distribution was balanced. 

\begin{figure*}[ht!]
\centering
\includegraphics[width=0.7\linewidth]{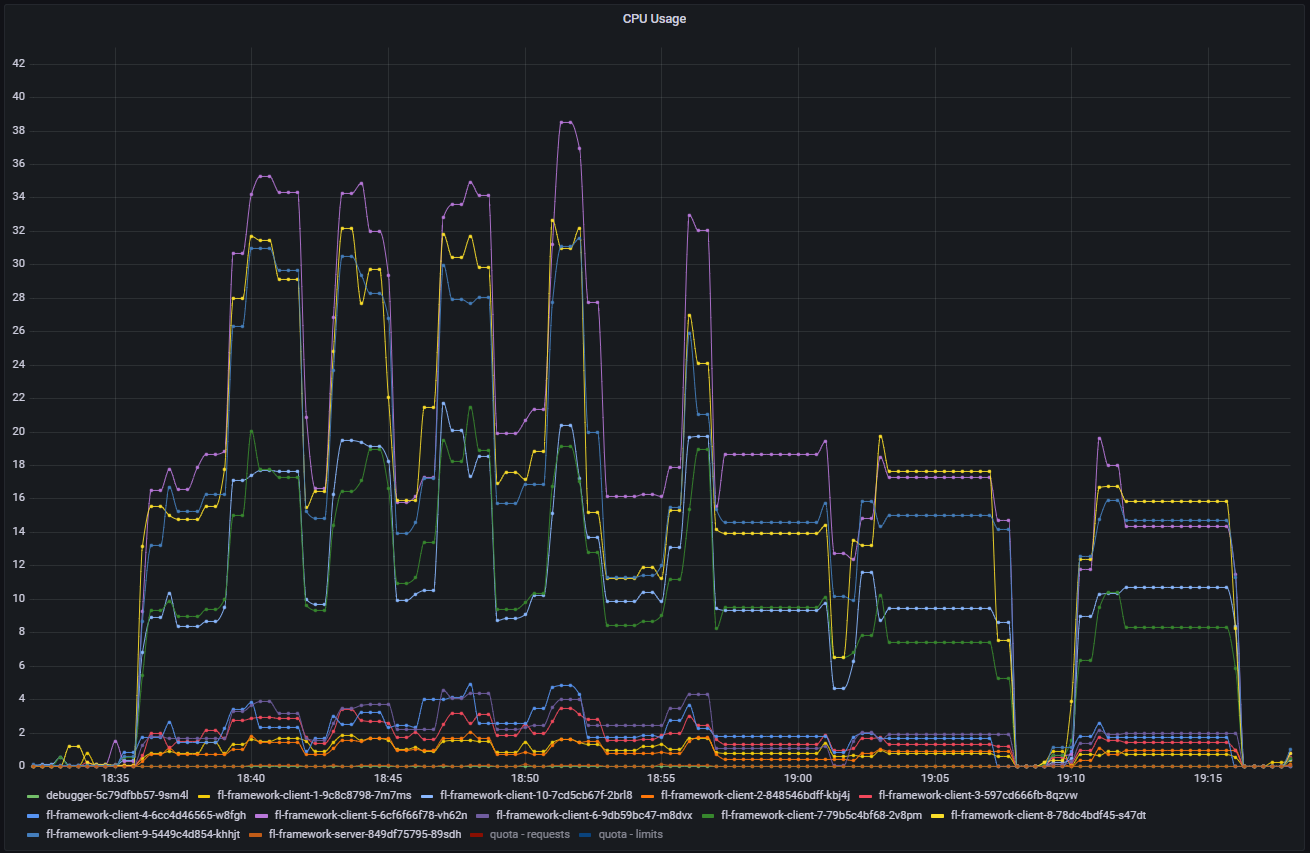}
\caption{CPU Usage from server and clients in experiment 1}
\label{fig:cpu_all_1}
\end{figure*}

The choice of using Grafana for experiment data visualization allows us to dynamically select which data points to plot in the chart. So, to better analyze the assumption about the first chart, we can choose some clients of specific hardware and others of another specification to compare them. Figure \ref{fig:cpu_different_3} demonstrates this selection using the same chart with now only client-5, client-4, and the server CPU usage being plotted. It becomes evident that client-4 (Intel Core i5-3210M CPU @ 2.50GHz) presented higher CPU resource usage and needed more time to process the same number of samples as client-5 (Intel Core i7-5500U CPU @ 2.40GHz). 

As for the server (orange line in Figure \ref{fig:cpu_different_3}), the peak of CPU usage is noticeable at the beginning of the experiment as a brief bump (near 1\%) before the first training round. This can be explained by the fact that the data distribution created by the server is done at the beginning of the experiment. We can also see that the server's CPU resource usage is almost negligible compared to any client's. The server only uses the CPU in between training rounds to aggregate the parameters received from the clients, which is a pretty simple task compared to the training done by the clients, thus explaining this phenomenon. The hardware profile of the server node is the same as client-5 (Intel Core i7-5500U CPU @ 2.40GHz), thus the amount of CPU resources needed for client and server tasks can be analyzed and understood in perspective.

To demonstrate that the analysis done in experiment 3 could be done in other experiments, we present a sample chart from experiment 1. Figure \ref{fig:cpu_all_1} shows the same chart of figure \ref{fig:cpu_all_3} plotted with data from the first experiment. It is interesting to see how the data distribution affects the CPU usage of the clients. We can visualize the gaps in between them due to the size of the samples being different for each client, even for those running in similar hardware profiles. Further analysis can be done in the Grafana dashboard to understand the behavior of each one of the clients individually and in any time frame specified. We will not present an exhaustive use of interactions with Grafana charts in this paper for the sake of brevity.

\subsection{Network Traffic}\label{subsection-network-traffic}

Similar to the CPU usage charts presented in the previous subsection, the same type of analysis can be done to visualize the network traffic between the clients and the server. Prometheus gathers information on traffic flowing in and out of Pods, collecting metrics such as, packets in/out, bytes in/out, and errors/drops, whereas Grafana enables interactive visualization and filtering of these metrics in different types of charts. To demonstrate this we created a simple dashboard with line charts containing bytes in/out of all Pods participating in the experiments. Two of these charts are presented in the following with data from the third experiment over the same time frame previously used in Figures \ref{fig:cpu_all_3} and \ref{fig:cpu_different_3}.

\begin{figure*}[t]
    \centering
        \includegraphics[width=0.8\linewidth]{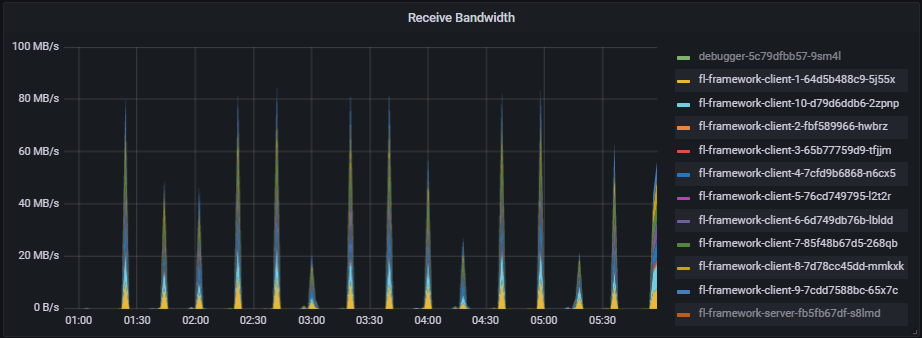}
    \caption{Receive bandwidth from server and clients in experiment 3}
    \label{fig:receive_all}
\end{figure*}

Figure \ref{fig:receive_all} shows the received (inward) traffic for each one of the clients involved in the third experiment as stacked areas. The X-axis indicates the timestamp of the metric, and the Y-axis indicates the network traffic in Megabytes per second (MBps). Each data sample is also collected with a 10-second interval between them. The first aspect to notice is that peaks of traffic are presented at the beginning and end of each round, which is expected since that is when communication between client and server should occur. The aggregated traffic peak considering all 10 clients was around 80 MBps. We did not show traffic sent (outward) from each client Pod because the communication follows a client-server paradigm, i.e., all the traffic flowing out of clients goes into the server and vice-versa. 

\begin{figure*}[t]
    \centering
        \includegraphics[width=0.8\linewidth]{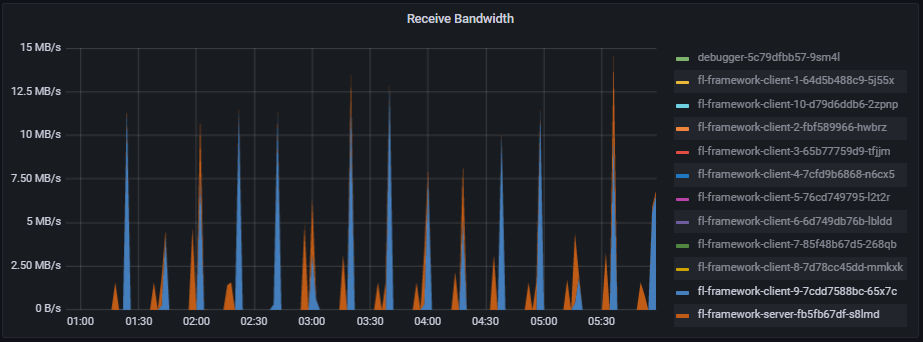}
    \caption{Receive bandwidth from server and client-9 in experiment 3}
    \label{fig:receive_server_client}
\end{figure*}

To better understand client and server behavior, we can filter only one client and the server in the same chart. For example, Figure \ref{fig:receive_server_client} shows the traffic received (inward) in the server and client-9 for the third experiment. One can see that the communication pattern is always the server receiving data and, afterward, the client. This is expected since, initially, the clients connect to the server, and the server accepts the connection. During training, the pattern is maintained since the server receives the parameters from the clients, and the client receives the aggregated parameters from the server. It is possible to visualize that inward traffic at the server side (received parameters from all clients) generates traffic peaks at most around 10-12 MBps, which means that in this experiment clients send less data than they receive. In other words, the amount of data required to represent the client parameters is smaller than the aggregated parameters generated at the server. This type of deeper understanding of the resource consumption patterns in the client-server interaction can help developers to better tune their FL-based systems.

The Prometheus monitoring application collects many other metrics, such as network packets, memory usage, or any other custom metric from applications configured in the system. For the purposes of demonstration, in this work, we considered only the CPU and the network due to their relevance in FL. Nevertheless, the interested experimenter can easily create new dashboards to monitor their experiment metrics in Grafana as they wish.
\section{Conclusion}\label{section-conclusion}

FL is an machine learning paradigm that is constantly being adopted due to its distributed approach, the necessity of data privacy, and the vast increase in IoT devices. Heterogeneous data distributions/contents of clients proved to be a real challenge and gained great notoriety over the last few years. Proposals, such as DRL algorithms to dynamically learn the weight of the contributions of each client at each round, continue to be developed and will be fundamental to increasing the accuracy and usability of FL in more applications. Testing and assessing this FL algorithm can be a challenging and complex task due to the systems' distributed nature and the possible scenarios.

To address this complexity, this work proposed a conceptual framework to facilitate testing FL scenarios in distributed computing environments with different types of data distributions. This work achieves the intended proposal by proposing a conceptual framework for testing FL scenarios and demonstrating an implementation of those concepts in the PoC solution. The solution developed shows that creating an edge-like FL testing framework that can scale to several different types of real-life scenarios using distributed heterogeneous computing and other data distributions is possible, inspiring further development of the concepts and improvement of the PoC solution.

To illustrate the capabilities of the PoC solution, three experiments with three different FL scenarios were conducted. The results showed how it is possible to analyze the impacts of class and data imbalance in a real-life distributed system of FL through the framework via the f1-measures outputted in the experiment results. It was also possible to see resource usage of applications via the monitoring solution and to demonstrate the impact of the underlying heterogeneous infrastructure used.  




The critical point taken from this work is that designing a solution for testing FL algorithms that has independence between infrastructure and applications can be very efficient and effective during the development phase. Containers enable reuse, isolation, and easy testing of applications regardless of the environment in which they were deployed, either locally or in a distributed cluster. This approach allows the system to scale horizontally by adding more nodes to the cluster or creating more application replicas. Therefore, by offering dynamic scalability and independence across layers, the proposed framework facilitates the configuration and testing of FL algorithms in heterogeneous environments. 

For example, users can modify various parameters, such as data distribution and global model aggregations, without requiring significant changes in the underlying infrastructure. The design flexibility, paired with tools like Prometheus and Grafana for resource monitoring, offers efficient evaluation of FL scenarios in distributed edge environments. Additionally, the framework's independence between the infrastructure and application layers, enabled by on-premises or cloud-based container orchestration systems like Kubernetes, provides high flexibility and isolation across deployments.

However, further improvement could be made to generate automated visualizations of each client and the server's aggregated results shown in Section~\ref{section-results}. For instance, it should be possible to output the results of the FL algorithm's performance to the Prometheus server so that it can be further visualized by Grafana or another visualization tool with access to its database. Moreover, fault tolerance experiments can also be done in the PoC solution to enable further improvement in the framework's reliability. From an application perspective, one way would be to test limited connectivity scenarios by disconnecting clients between training, a common scenario in edge environments. For this purpose, cloud-native chaos engineering tools, such as Chaos Mesh\footnote{https://chaos-mesh.org/ (accessed October 8th, 2024)}, could be integrated into the framework to simulate abnormalities, allowing users to find and prevent potential failures in production environments. 

Stress testing would also help understand the platform's limitations, as some results indicate performance bottlenecks in cases of non-IID data distribution. For instance, in more complex classification problems like those in Experiment 3 using the CIFAR-100 dataset, the framework struggled to maintain model performance across clients with limited data samples per class. This observation suggests that further evaluations, particularly addressing class imbalance and heterogeneity in data distribution, can better demonstrate the framework's capability to deal with more challenging scenarios.

Still, the framework's flexibility and scalability make it a promising option for testing FL algorithms in edge-like environments, opening new opportunities to develop experiments with them. All of the code is publicly available and has been developed with extensibility in mind. Future works include extending the datasets, models, and FL strategies supported and improving the applications as a whole.

\section*{Acknowledgments}
This research is part of the \textit{INCT of Intelligent Communications Networks and the Internet of Things (ICoNIoT)} funded by CNPq Process 405940/2022-0 and CAPES Finance Code 88887.954253/2024-00. This study was also partially funded by CAPES Finance Code 001 and FAPESP \textit{Distributed intelligence in communications networks and in the internet of things (IDRIC)} project grant 23/00673-7.



 \bibliographystyle{elsarticle-num} 
 \bibliography{cas-refs}





\end{document}